\documentclass[10pt,journal,compsoc]{IEEEtran}
\newif\ifpeerreview

\peerreviewfalse

\usepackage[nocompress]{cite}
\usepackage{url}
\usepackage{amsmath,amssymb,graphicx}
\graphicspath{{../}}
\usepackage{booktabs}
\usepackage{subcaption}
\usepackage{float}
\usepackage{epigraph}
\usepackage{microtype}
\usepackage{multirow}
\usepackage{xspace}
\usepackage{bm}
\usepackage{booktabs}

\usepackage{lipsum} \usepackage[pagebackref,breaklinks,colorlinks]{hyperref}

\usepackage{color}

\newcommand{\ie}{\textit{i.e}\xspace}
\newcommand{\eg}{\textit{e.g}\xspace}
\newcommand{\etal}{\textit{et al.}\xspace}
\DeclareMathOperator*{\argmax}{arg\,max}

\usepackage[capitalize]{cleveref}
\crefname{section}{Sec.}{Secs.}
\Crefname{section}{Section}{Sections}
\Crefname{table}{Table}{Tables}
\crefname{table}{Tab.}{Tabs.}

\newcommand{\squishlist}{
	\begin{list}{$\bullet$}
		{ \setlength{\itemsep}{0pt}
			\setlength{\parsep}{1pt}
			\setlength{\topsep}{1pt}
			\setlength{\partopsep}{0pt}
			\setlength{\leftmargin}{1.5em}
			\setlength{\labelwidth}{1em}
			\setlength{\labelsep}{0.5em} } }
\newcommand{\squishend}{\end{list} 
}

\usepackage[switch]{lineno}

\newcommand{\paperID}{4}

\title{Robust Scene Inference\\under Noise-Blur Dual Corruptions}

\author{Bhavya Goyal, Jean-François Lalonde, Yin Li, Mohit Gupta
\IEEEcompsocitemizethanks{\IEEEcompsocthanksitem B. Goyal, Y. Li and M. Gupta are with University of Wisconsin-Madison, Madison, WI, 53715.\protect\\
E-mail: bhavya@cs.wisc.edu
\IEEEcompsocthanksitem J. Lalonde is with Université Laval, Québec, QC, Canada
}}

\begin{document}

\IEEEtitleabstractindextext{\begin{abstract}
Scene inference under low-light is a challenging problem due to severe noise in the captured images. One way to reduce noise is to use longer exposure during the capture. However, in the presence of motion (scene or camera motion), longer exposures lead to motion blur, resulting in loss of image information. This creates a trade-off between these two kinds of image degradations: motion blur (due to long exposure) vs. noise (due to short exposure), also referred as a dual image corruption pair in this paper. With the rise of cameras capable of capturing multiple exposures of the same scene simultaneously, it is possible to overcome this trade-off.
Our key observation is that although the amount and nature of degradation varies for these different image captures, the semantic content remains the same across all images. To this end, we propose a method to leverage these multi exposure captures for robust inference under low-light and motion. Our method builds on a feature consistency loss to encourage similar results from these individual captures, and uses the ensemble of their final predictions for robust visual recognition. We demonstrate the effectiveness of our approach on simulated images as well as real captures with multiple exposures, and across the tasks of object detection and image classification.
Project: \url{https://wisionlab.com/project/noiseblurdual}
\end{abstract}

}

\maketitle
\thispagestyle{empty}

\IEEEraisesectionheading{
  \section{Introduction}\label{sec:intro}
}

\IEEEPARstart{I}maging trade-offs are a fundamental characteristic of any computer vision system, and they are often exacerbated when the imaging conditions are challenging.
One such challenging condition is low-light and motion (scene or camera).
In such conditions, various types of corruptions are bound to be present in the image, and one can only compromise between them without ever removing them completely.
For example, low light could cause the images captured by the camera to exhibit strong noise.
While it is possible to mitigate noise by capturing longer exposures (or larger apertures), this often results in strong motion (or defocus) blur, leading to another kind of image quality degradation.
Hence, noise and blur represent ``dual corruptions''---reducing one (\eg, by adjusting the exposure) necessarily increases the other. 

\begin{figure*}
\centering
\begin{subfigure}{0.24\linewidth}
    \centering
    \includegraphics[width=\linewidth]{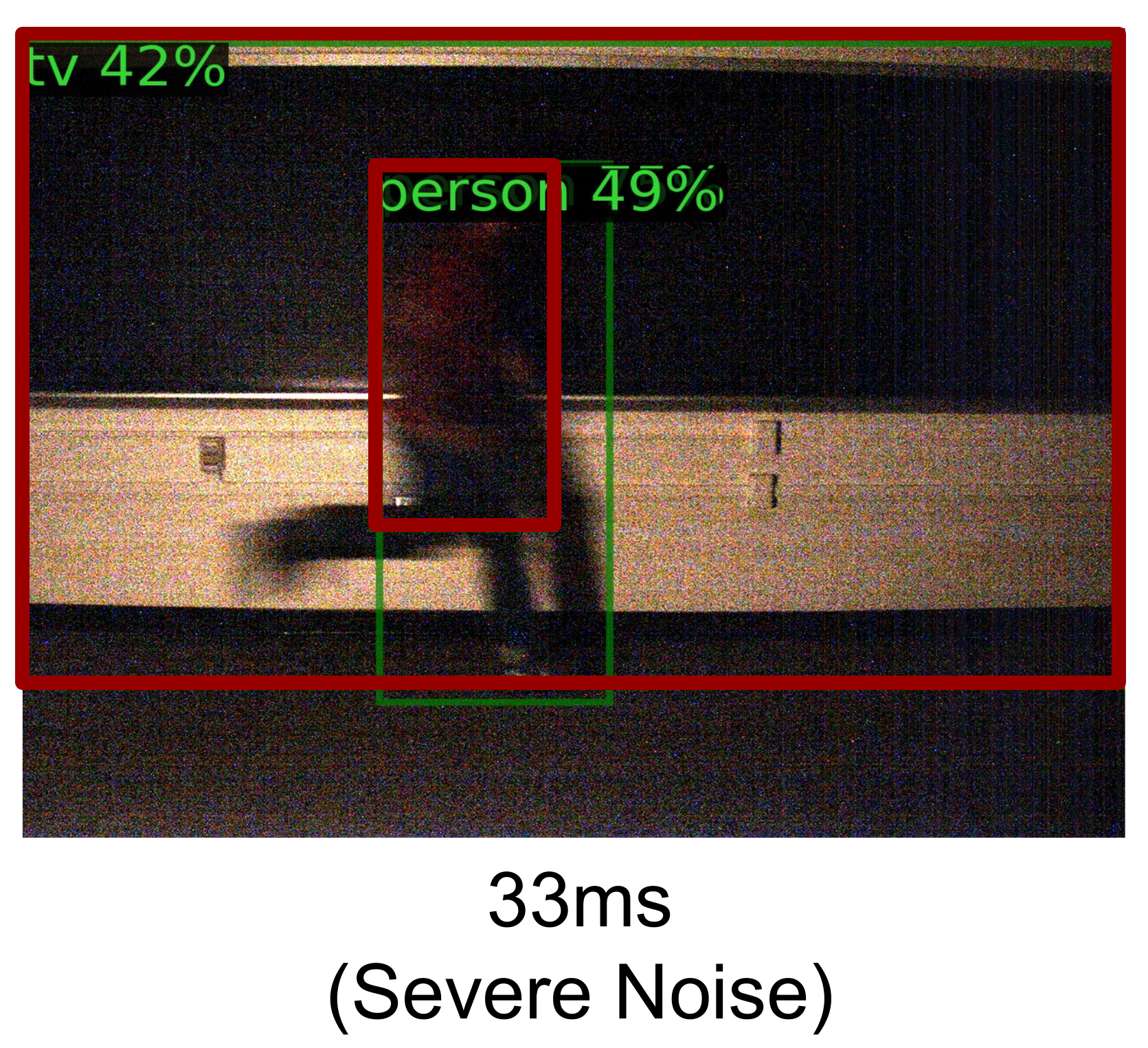}
    \caption{}
    \label{fig:teasera}
\end{subfigure}
\begin{subfigure}{0.24\linewidth}
    \centering
    \includegraphics[width=\linewidth]{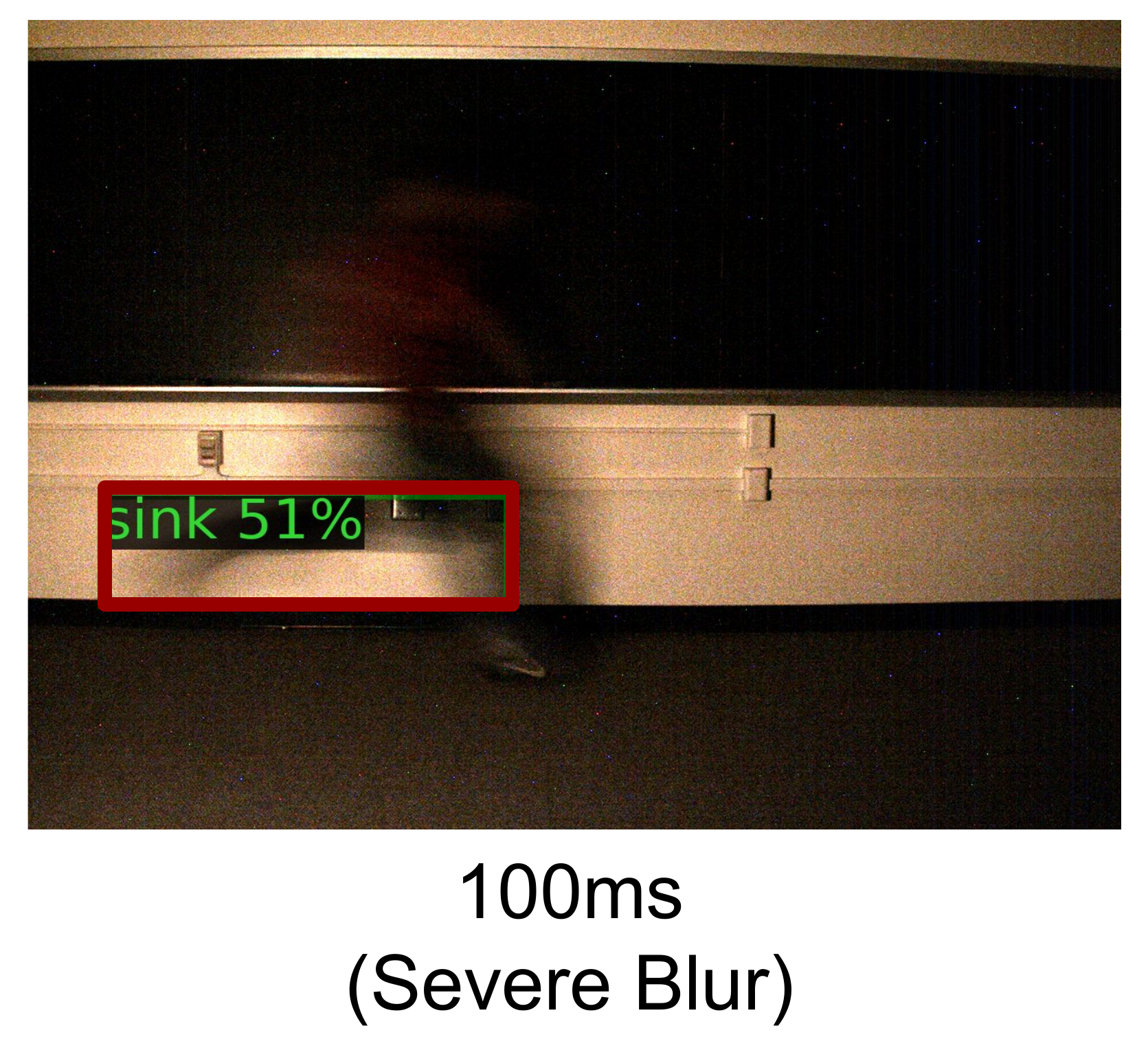}
    \caption{}
    \label{fig:teaserb}
\end{subfigure}
\begin{subfigure}{0.24\linewidth}
    \centering
    \includegraphics[width=\linewidth]{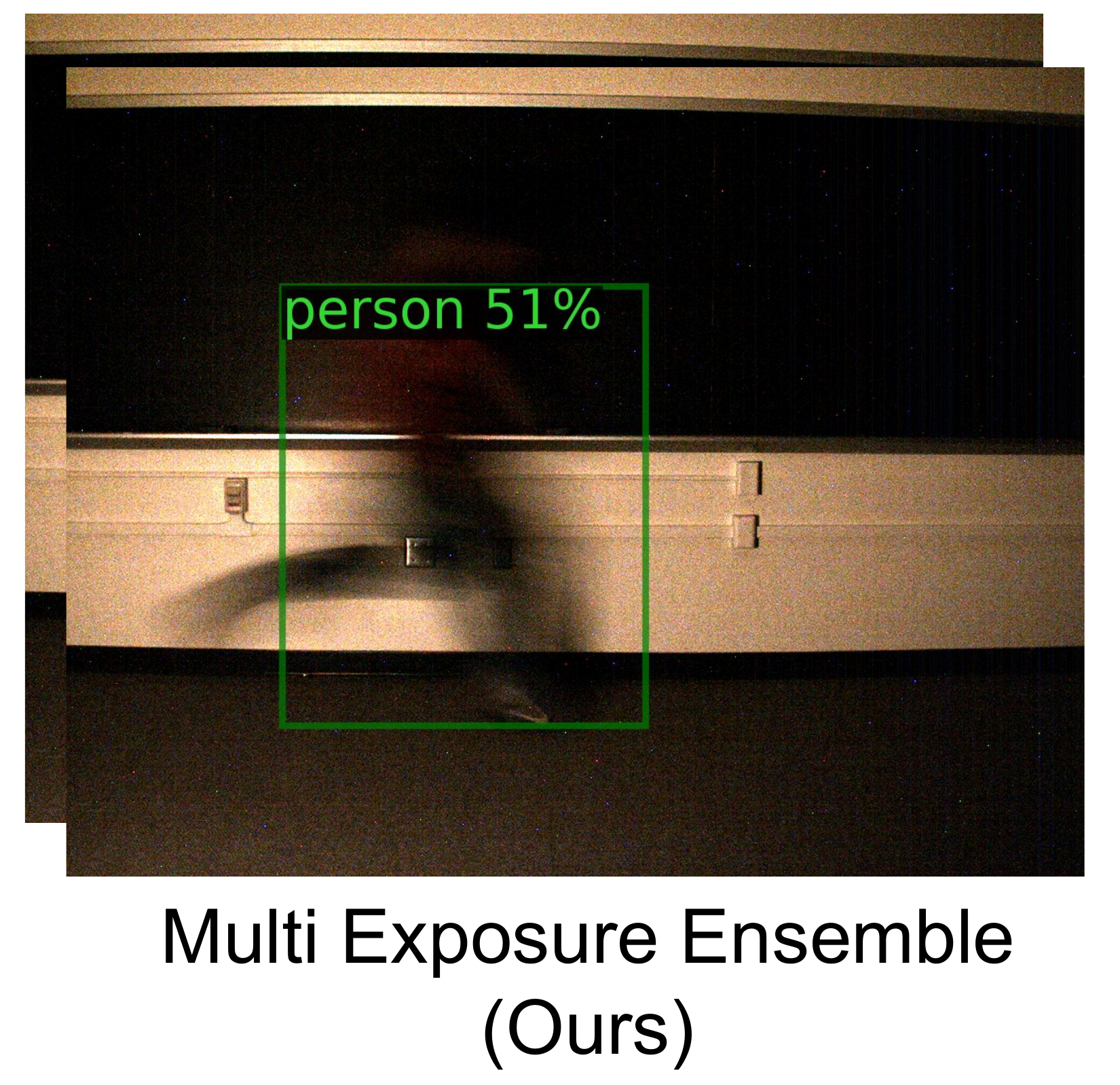}
    \caption{}
    \label{fig:teaserc}
\end{subfigure}
\begin{subfigure}{0.24\linewidth}
    \centering
    \includegraphics[width=\linewidth]{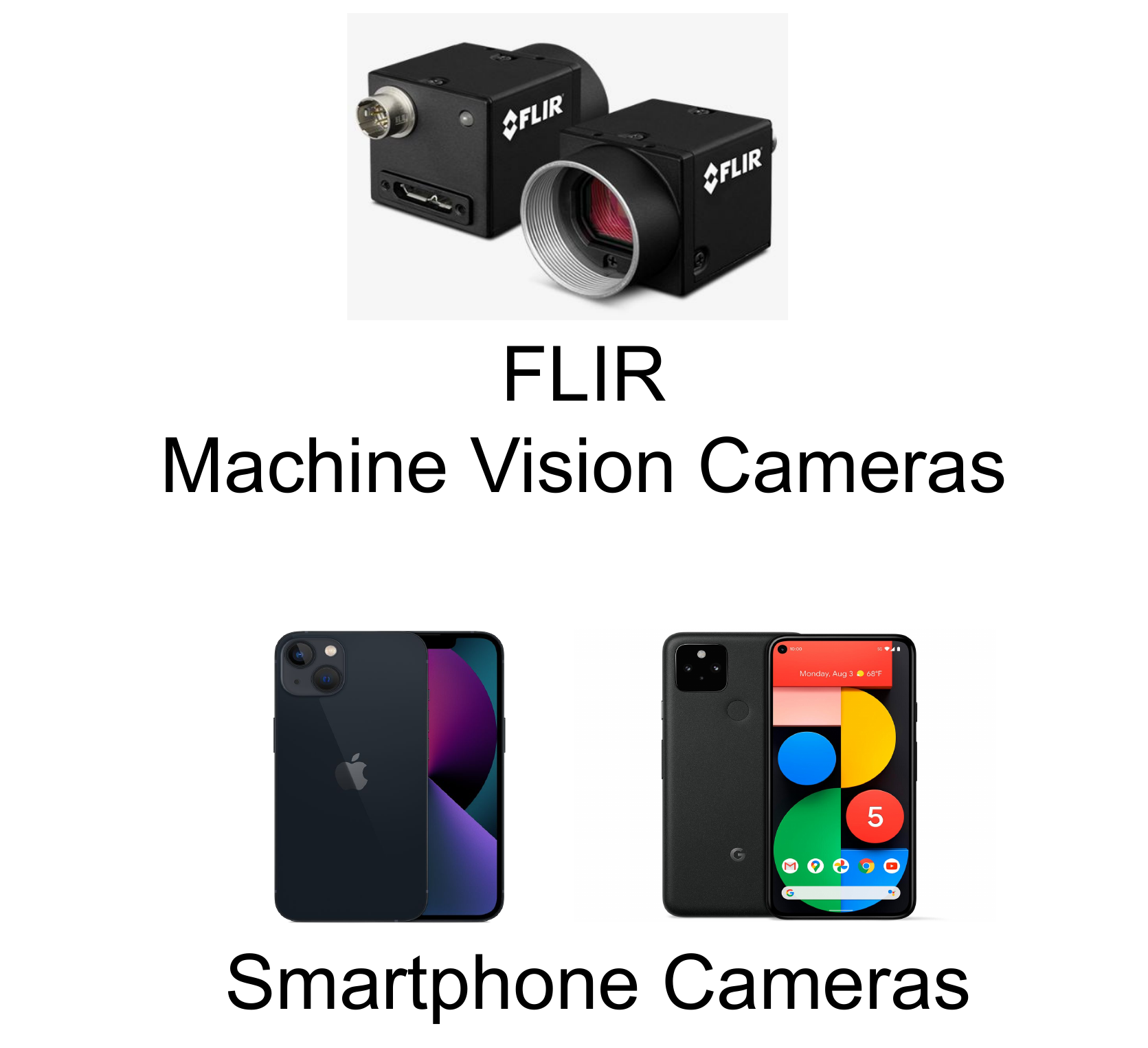}
    \caption{}
    \label{fig:teaserd}
\end{subfigure}
\vspace{-5pt}
\caption{\textbf{Multi Exposure Ensemble:} Figure shows a scene containing a fast moving object under low-light. Images with short exposure (\subref{fig:teasera}) and long exposure (\subref{fig:teaserb}) suffer from dual corruption: noise and/or blur.
Inference tasks like object detection on these images are severely affected: numerous false positives and wrong bounding boxes. (\subref{fig:teaserc}) Our approach leverages multiple captures of varying exposures for robust inference: accurate and tight bounding boxes.
(\subref{fig:teaserd}) Such multi-exposure images are easy to capture with machine vision cameras or modern smart phone cameras (eg. Google pixel and iPhone) that uses burst photography for HDR imaging.
}
\label{fig:teaserexample}
\end{figure*}

\epigraph{\emph{All happy families are alike; \\ each unhappy family is unhappy in its own way.}}{Leo Tolstoy}

Consider a set of images captured under low-light at varying exposures (Figure~\ref{fig:teaserexample}), thereby spanning the space of noise-blur ``dual corruptions''. Each image, being corrupted in its own way, offers a different ``window'' on the scene: moving objects will appear sharper when the exposure is lower, while static low-contrast regions will be more easily perceptible in longer exposures. In order words, while any single image from the set might never be optimal in challenging scenarios, the \emph{set of images spanning the dual corruption space} contains much richer and complementary information that can be leveraged for performing robust scene inference even under challenging imaging scenarios.

In this paper, we propose the idea of performing scene inference in the space of noise-blur corruption. Our \emph{key observation} is utilizing the ``persistence of prediction'' across differently degraded images of the same scene, significantly higher accuracy can be achieved as compared to performing inference on individual images. Figure~\ref{fig:teaserexample} shows an example. Although differently degraded images have different low-level features, the semantic content remains the same across all images. We develop techniques that encourage similar predictions from individual captures, and aggregate the predictions across individual images for robust visual recognition.  

We demonstrate the proposed approaches on two visual recognition tasks, namely image classification and object detection. We perform experiments on large scale datasets of real images with synthetic corruptions and show that performing inference on a set of dual corruption images outperforms conventional baselines in extreme low-light and motion conditions. Finally, we also show improved performance on real-world experiments using machine vision sensors. \medskip

\noindent {\bf Scope and Limitations:} While implementing this idea requires capturing multiple exposures, most modern cameras already allow varying imaging parameters (\eg, exposure, aperture) in rapid succession. For example, modern cell phone cameras can take multiple snaps with a variety of exposures and fuse them to create an aesthetically pleasing image~\cite{hasinoff2016burst}. Increasingly, machine vision sensors~\cite{commaai}
are also starting to perform exposure bracketing to capture high dynamic range (HDR) images for autonomous driver assist systems, while others go further and offer the capability of \emph{simultaneously} capturing different exposure images via a spatially varying exposure sensor for HDR imaging~\cite{Nayar:2000} and motion-deblurring~\cite{nguyen2022learning}. These ongoing developments in camera technology, coupled with the proposed computational techniques can lead to the next generation of computer vision systems which will perform reliably even in non-ideal real-world scenarios (\eg, imagine an autonomous car driving on a dark night attempting to detect pedestrians) where it is extremely challenging for conventional algorithms to extract meaningful information reliably. 

 \section{Related Work}
\label{sec:related}

\noindent \textbf{Image Corruptions and Benchmarks}. There has been some recent interest in simulating common image corruptions and benchmarking their adversarial effect on the performance of computer vision models, especially those relying on deep models~\cite{hendrycks2018benchmarking,michaelis2019benchmarking}. In parallel, developing robust visual inference methods has also received much attention. For example, a teacher-student framework was proposed~\cite{xie2020self} to improve image classification performance. Several noise and corruption models have been considered, including both physics-based~\cite{wei2020physics} and learning-based~\cite{abdelhamed2019noise}. Efforts in capturing real datasets of noisy images have also been pursued. A dataset of images captured in low light with annotations for object detection~\cite{loh2019getting} has been collected. Another examples is the dataset containing low-light and corresponding well-lit cellphone images for denoising~\cite{abdelhamed2018high}, which has recently been extended to videos in~\cite{wang2021seeing}. Most previous works simulate or collect real captures with image degradations like noise in low-light, but we consider a more challenging and practical setting where both low-light and motion are presented, and hence dual image degradation come into play.

\medskip
\noindent \textbf{Noise Removal and Deblurring}.
Due to its importance in image processing, denoising and/or deblurring degraded images has been a very popular topic for decades. Recently, numerous works have been proposed using neural networks for deblurring~\cite{xu2014deep, schuler2015learning, zhang2017learning} and denoising~\cite{zhang2017beyond}. For example, sparse denoising auto-encoder was considered for robust denoising~\cite{agostinelli2013adaptive}. A recent line of work proposes to perform joint denoising and inference on noisy images~\cite{liu2019transferable, liu2020connecting, diamond2017dirty}. 
While existing image restoration methods can obtain high quality reconstructions, performing inference directly on the corrupted images does not require any pre-processing and is thus more efficient and as we demonstrate, can achieve increased robustness under severe image degradation. Alternatively, other methods aim to design cameras that produce better images directly, either by optimizing the hyperparameters of existing image signal processors (ISP)~\cite{tseng2019hyperparameter} or, by designing novel ISPs~\cite{Heide2014FlexISP,Gharbi2016DeepJD,Chen2017FastIP,chen2018learning}. These methods may, however, not entirely remove noise in challenging low-light situations, due to the fundamental limitation of the optics and sensors.

\medskip
\noindent \textbf{Inference on Corrupted Images}. Many recent works tackle different inference tasks \emph{directly} on images with common corruptions. Rozumnyi~\etal \cite{Rozumnyi_2021_ICCV} proposed a matting and deblurring network for faster inference for the detection of fast moving objects in videos. Cui~\etal \cite{cui2021multitask} designed a multitask auto-encoder for image enhancement, which leverages a physical noise model and ISP setting in a self-supervised manner to improve detection performance. Wang \etal \cite{wang2021regularizing} presented a framework for monocular depth estimation under low-light using self-supervised learning and demonstrate their results on nighttime datasets. Others have used knowledge distillation techniques for image classification under low-light~\cite{gnanasambandam2020image}, or for object detection by leveraging bursts of short exposure frames~\cite{Li_2021_ICCV}. Goyal~\etal \cite{goyal2021photon} used a single photon camera and proposed to train on a wide spectrum of images at various SNR, with encouraging results on image classification and monocular depth estimation.
Song~\etal \cite{song2021matching} introduced a technique for image matching using local descriptors and initial point-matching methods for extremely low-light images in RAW format.
Wang~\etal \cite{wang2020deep} proposed to learn the mapping relationship between representations of low and high quality images, and used it as a deep degradation prior (DDP) for image classification on degraded images. Adversarial Logit Pairing~\cite{kannan2018adversarial} also provides some robustness to the inference on noise and blur corruptions~\cite{hendrycks2018benchmarking} by matching logits output of clean image with adversarial perturbed image.

Our goal is different from all previous approaches. We propose techniques that leverage the space of noise-blur dual corruptions rather than looking at a single image corruption. We show that our approach is versatile for several downstream tasks, including image classification and object detection.

\medskip
\noindent \textbf{Leveraging Multiple Captures}.
Multiple exposures can be used to reconstruct high dynamic range (HDR) images~\cite{debevec-siggraph-97}, even in the presence of motion~\cite{sen2012robust,kalantari2019deep}. Hasinoff~\etal~\cite{hasinoff2009time,hasinoff2010noise} proposed ways to select settings for these multiple captures like ISOs and focus settings. The popularity of mobile photography has led to the further development of burst photography~\cite{hasinoff2016burst}, which has been used for denoising~\cite{mildenhall2018burst}, deblurring~\cite{delbracio2015burst,aittala2018burst}, and super-resolution~\cite{wronski2019handheld}. In sharp contrast, we exploit multiple exposures for high-level inference tasks such as classification and detection, rather than low-level image reconstruction.

 \begin{figure*}
\begin{center}
\includegraphics[width=\linewidth]{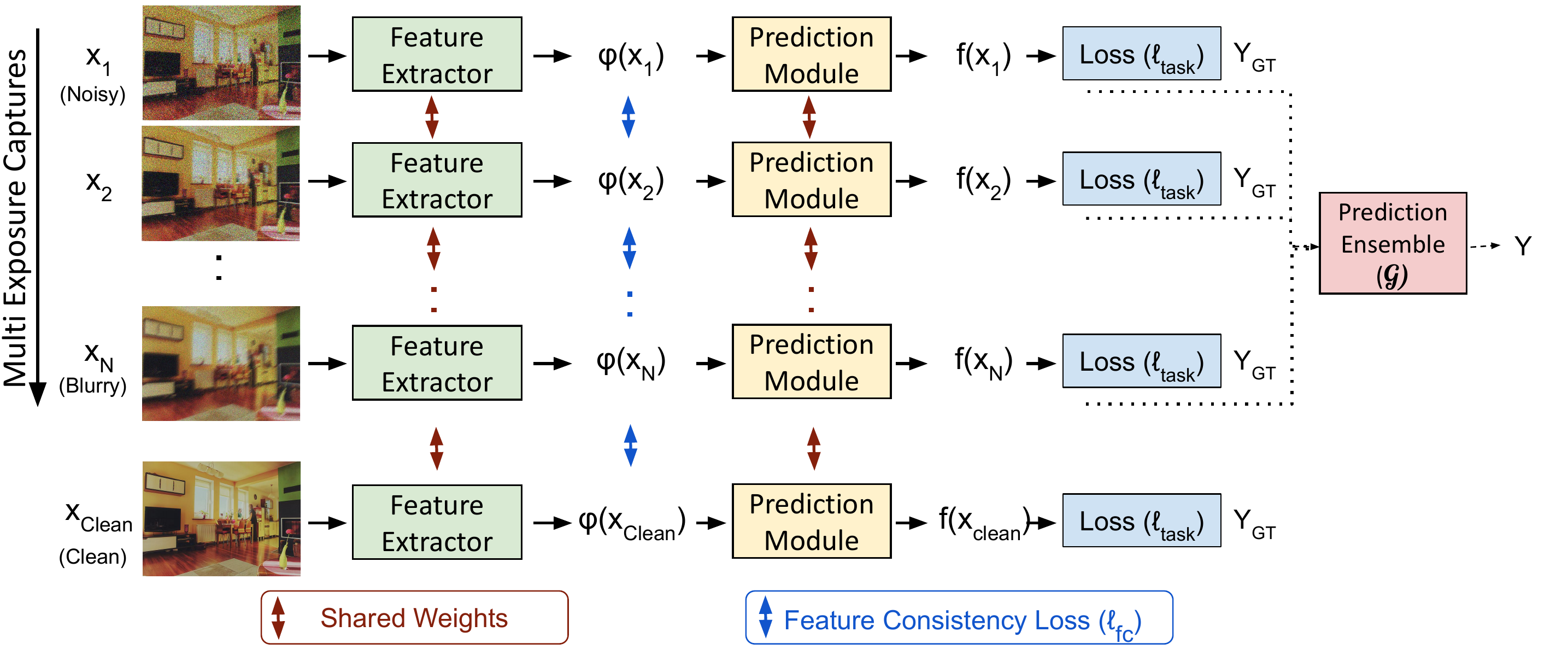}
\end{center}
\vspace{-5pt}
\caption{\textbf{Architecture Overview:} Our approach trains an inference model using multiple captures of varying exposures, all containing the same semantic content but different amounts of noise-blur dual corruptions.
We introduce feature consistency loss during training to enforce consistency of feature outputs from each individual captures. During testing (dashed lines), our model returns the ensemble prediction using each individual capture to produce final output for more robust prediction.
}
\label{fig:arch}
\end{figure*}

\section{Dual Corruption Space}
\label{sec:duality}

\noindent \textbf{Noise Blur Trade-off in Image Formation.}
We consider the relationship of noise and blur with the exposure time under low-light conditions and in the presence of scene (or camera) motion. The number of photons incident at a given pixel during a short exposure time is small under low-light conditions. Because of this, noise becomes dominant in the captured images and has to be properly modeled. 

In the presence of scene/camera motion, let the photon flux (photons/second) at a pixel $p$ on time $t$ be $\phi_{p,t}$. The key is to consider that the incident flux at each pixel changes over time $t$, since the pixel may image different scene points due to scene/camera motion, resulting in an image $x$ with motion blur. Assuming an exposure time $\Delta t$ and a linear camera with quantum efficiency $\eta$, the raw reading at pixel $p$ (without quantization) is given by
\begin{equation}
I_p =  \int_0^{\Delta t} \phi_{p, t} \eta \ dt + z_p
\end{equation}
where $z_p$ is the noise at pixel $p$. Here we ignore the non-uniformity of photon response and noise~\cite{granados2010optimal}, and consider three sources of noise.

\squishlist
\item \emph{Shot noise} $z^s_p$ refers to the inherent natural variation of the incident photons due the Poisson process of photon arrival $\mathcal{P}$ and is modelled as the square root of the signal. Therefore, $z^s_p \sim \mathcal{P}\left(\int_0^{\Delta t} \phi_{p,t}\eta \,dt \right)$. 

\item \emph{Readout noise} $z^r_p$ comes from the process of quantizing the electronic signal as well as electrical circuit noise, which is modelled as a zero mean Gaussian with variance $\sigma_r^2$ at each readout. Namely, 
$z^r_p \sim \mathcal{N}(0, \sigma_r^2)$. 

\item \emph{Dark current} $z^d_p$ arises due to thermally generated electrons and also follows a square root relationship with signal with a variance of $\sigma_d$. We thus have $z^d_p \sim \mathcal{P}(\sigma_d \Delta t)$.
\squishend\smallskip

We further assume that $z^s_p$, $z^r_p$, and $z^d_p$ are independent of each other, and follow an additive noise model~\cite{Hasinoff2010}, such that $z_p = z^s_p + z^r_p + z^d_p$~\cite{granados2010optimal}. Thus, $\mathrm{Var}(z_p) = \mathrm{Var}(z^s_p) + \mathrm{Var}(z^r_p) + \mathrm{Var}(z^d_p)$. This leads to the derivation of the signal-to-noise ratio (SNR) for the captured images, given by
\begin{equation}
\textnormal{SNR} = \frac{\left(\int_0^{\Delta t} \phi_{p, t} \eta \ dt\right)^2}{\int_0^{\Delta t} \phi_{p, t} \eta \ dt + \sigma_r^2 + \sigma_d \Delta t} \,.
\label{eq:snr}
\end{equation}
Under the presence of both low-light and motion, longer exposure time leads to improved SNR, as the noise increases slower than the signal. This, however, comes at a cost of increased motion blur in the captured images due to the integral of the incoming flux $\phi_{p,t}$. Hence, the exposure time allows us to trade off noise and blur in the image degradation space, which we term as \textit{Dual Corruption Space}.

\medskip
\noindent \textbf{Dual Corruption}. Our key idea is to leverage the spectrum of dual-corruption images by varying the camera parameters, resulting in a set $\mathcal{I} = \{x_1, ... x_N\}$ of images with different low-level characteristics (\eg, different amounts of blur and noise). For example, varying exposure time $\Delta t$ creates a sequence of images where noise gradually decreases but the amount of blur increases. An example such sequence is shown in Figure~\ref{fig:teaserexample}. Since these images are captured simultaneously (or in rapid succession), we can assume that they have similar semantic content.

\medskip
\noindent \textbf{An Image without Noise and Blur}. A special and theoretically interesting case in the dual corruption space is an \emph{ideal clean image} $x_{clean}$ captured using a very short exposure time ($\Delta t \rightarrow 0$) and without noise corruption ($z_p$ = 0). Such an image is free of noise and blur. Despite physically implausible, this construct is sometimes convenient for our derivations.

\section{Scene Inference under Noise-Blur Dual Corruptions}

We consider scene inference tasks represented as an inference module $f(x) \equiv g \circ \varphi(x)$, where, without loss of generality, $\varphi(x)$ is a feature extractor, and $g$ is a prediction module. Here, $\circ$ is the composition operator. $f(x)$, oftentimes represented by a neural network, maps an input image $x$ into its semantic label $y$. This generic formulation covers several vision recognition tasks, including image classification where $y$ is a categorical label, and object detection where $y$ is a set of labeled bounding boxes. We further assume that this function $f(\cdot)$ is learned from data by minimizing a certain loss function. 

Given a set of $N$ noise-blur dual corruption images $\mathcal{X} = \{x_1, ... x_N\}$ capturing the same scene, our key intuition is that despite differences in low-level \emph{image} features (\eg, pixel values), their \emph{latent} features should remain similar. In what follows we formulate this intuition as a data prior, devise the training and inference schemes, and demonstrate interesting properties of the resulting method.

\subsection{Robust Inference with Multiple Exposures}
\label{sec:robustinference}
A simple prior is to assume that the latent features $\{\varphi(x_1), ... \varphi(x)_N\}$ follow a Gaussian distribution, centered at the ideal clean image $x_{clean}$ and with a small variance $\epsilon^2$. This prior ensures that with high probability the distance between any pair of latent features will stay in a small $\ell_2$ radius controlled by $\epsilon^2$. With such assumption, we arrive at the following conditional probability $p(y | x)$ for scene inference. 
\begin{equation}
p(y | x) \propto p(y | \varphi(x)) p(\varphi(x) | x),
\label{eq:cond_prob}
\end{equation}
where $p(y | \varphi(x))$ is given by the prediction module $g$, and $p(\varphi(x) | x) \sim \mathcal{N}(\varphi(x_{clean}), \epsilon^2)$ represents the data prior. We now describe the training and inference schemes based on this formulation, as illustrated in Figure~\ref{fig:arch}.

\medskip
\noindent {\bf Training with Multiple Exposures}. Given the ground-truth label $y$, minimizing the negative log likelihood of Equation~\ref{eq:cond_prob} on a training sample (a set of images $\{x_i\}$ spanning the dual corruption space) leads to the following loss function 
\begin{equation}
\small
    \ell = \sum_i^N \ell_{task}(p(y | \varphi(x_i)), y) + \frac{1}{\epsilon^2} \sum_i^N \| \varphi(x_i) - \varphi(x_{clean}) \|_2^2.
\label{eq:loss}
\end{equation}
Here, we slightly abuse the notation to replace the first term $-\log(p(y | \varphi(x)), y)$ with a more general task-specific loss $\ell_{task}(p(y | \varphi(x)), y)$. It is easier to consider the case of image classification, where the target $y$ is a categorical variable. The term of $-\log(p(y | \varphi(x)), y)$ becomes the cross-entropy loss, commonly used for classification. When $y$ moves beyond simple categorical or scalar outputs (\eg, for the object detection task), Equation~\ref{eq:loss} allows to plug in any loss function $\ell_{task}$ tailored for the task. On the other hand, the second term can be interpreted as a feature consistency loss, re-weighted by a coefficient as the reciprocal of the Gaussian variance ($1/\epsilon^2$).

Our loss function in Equation~\ref{eq:loss} assumes that a reference clean image is available during training, as often the case in our experiments. When such a clean image is not presented, we simply replace the second term with its equivalent form that only involves the summation of pairwise distances between $\varphi(x_i)$ and $\varphi(x_j)$, \ie, $\frac{1}{2N\epsilon^2}\sum_{i, j} \|\varphi(x_i) - \varphi(x_j)\|_2^2$.

\medskip
\noindent {\bf Inference with Model Aggregation}.
At inference time, the maximum likelihood estimation of Equation~\ref{eq:cond_prob} is not viable without the clean image $x_{clean}$. Instead, we resort to using the ensemble of the predictions from individual multi-exposure images as the final output prediction. Our key intuition is that no individual capture in the dual corruption space captures all the necessary information that may be required for the robust inference, but the ensemble output is more effective as it uses the predictions from individual images that contribute with the relevant information individually. This is given by
\begin{equation}
f(\mathcal{X}) = \mathcal{G}(f(x_1), f(x_2)...f(x_N)) \,,
\label{eq:outputensemble}
\end{equation}
where $\mathcal{G}$ is an aggregate function to get the ensemble prediction. $\mathcal{G}$ is highly flexible and often task-relevant. For example, for the image classification task, $\mathcal{G}$ could be a simple average operator over the probability outputs. For object detection, $\mathcal{G}$ might be a voting scheme of detected objects. By aggregating multiple model outputs, Equation~\ref{eq:outputensemble} is conceptually similar to the well-known model ensemble~\cite{rokach2010ensemble}.

\medskip
\noindent {\bf Certified Robustness}.
When considering a classification problem with $c$ categories (\eg, image classification), we notice an interesting link between our inference scheme and a well-known robust classifier~\cite{cohen2019certified}. Specifically, when $\mathcal{G}$ is an average operator and the decision is made by taking the category with the highest confidence from $f(\mathcal{X})$, our inference defined a ``smoothed'' classifier with certified robustness~\cite{cohen2019certified} under the Gaussian distribution
\begin{equation}
\begin{split}
& \argmax\ p(g(\hat{\varphi}(x))=c),\\
& \textnormal{where}\quad \hat{\varphi}(x) \sim \mathcal{N}(\varphi(x_{clean}), \epsilon^2).
\end{split}
\end{equation}
Cohen \etal~\cite{cohen2019certified} showed that such a classifier, if passes additional certification, is robust within a certain $\ell_2$ radius around $\varphi(x_{clean})$. Intuitively, this indicates that our model will produce consistent results (the same as ones given by the clean image) for all corrupted images spanning the dual corruption space, should the Gaussian assumption is satisfied. We deem theoretic investigation into this direction as our future work.

\section{Evaluation of Robust Scene Inference}

We demonstrate the effectiveness of our method on two important scene inference tasks: object detection and image classification.

\subsection{Object Detection}
\label{sec:detection}

\noindent \textbf{Instantiation}. Figure~\ref{fig:arch} shows the overview of our approach using multi-exposure ensemble for the object detection task. We implement our approach using the single-stage FCOS architecture~\cite{tian2019fcos}. The output prediction of the FCOS model for image of size $H \times W$ consists of pixel-wise classification scores ($H \times W \times C$) for $C$ object categories, centerness scores ($H \times W \times 1$) and bounding box coordinates regression outputs ($H \times W \times 4$). During inference, our ensemble predictor ($\mathcal{G}$), takes the pixel-wise classification scores, centerness scores and box coordinates, and returns their average at each FPN level. 
Loss function for the inference task ($\ell_{task}$) is the same as defined in FCOS architecture (i.e. sum of focal loss, regression loss for bounding boxes and centerness loss. Refer~\cite{tian2019fcos} for details).
Our feature consistency loss ($\ell_\mathrm{fc}$) is the L2 distance between feature outputs from the CNN network (final layer after global average pooling).

\begin{table*}[th!]
\centering
\begin{tabular}{l | c c c c c | c c c c c | c c c c}
\toprule
\multirow{2}{*}{Method} & \multicolumn{4}{c}{REDS} && \multicolumn{4}{c}{{CityScapes}} &&  \multicolumn{4}{c}{{MS-COCO}}\\
& mAP & APs & APm & APl && mAP & APs & APm & APl && mAP & APs & APm & APl\\
\midrule
Clean Model & 16.36 & 17.96 & 18.46 & 16.45 && 2.72 & 0.22 & 2.47 & 7.02 && 3.35 & 0.21 & 2.51 & 7.69   \\
Stylized Training~\cite{michaelis2019benchmarking} & 19.13 & 18.11 & 21.64 & 23.71  && 6.75 & 0.24 & 3.32 & 20.00 && 7.89 & 0.25 & 3.13 & 17.07 \\
Single Exposure & 30.17 & 20.27 & 25.75 & 36.88 && 18.07 & 3.96 & 15.77 & 35.54 && 21.25 & 6.58 & 22.39 & 33.88 \\
Denoising (BM3D)~\cite{dabov2007image} & 30.25 & 20.28 & 25.90 & 37.08 && 18.01 & 3.82 & 15.53 & 35.97 && 21.78 & 6.76 & 22.78 &  34.43\\
Denoising (MPRNet)~\cite{Zamir2021MPRNet} & 25.67 & 18.97 & 23.47 & 31.84 && 15.26 & 2.97 & 13.89 & 34.11 && 18.78 & 5.12 & 17.45 & 27.13 \\
Deblurring~\cite{carbajal2021non}  & 30.68 & 18.82 & 26.36 & 36.02 && 17.67 & 3.63 & 15.90 & 34.67 && 12.42 & 2.52 & 11.77 & 21.11\\
Denoising~\cite{dabov2007image} + Deblurring~\cite{carbajal2021non}  & 29.45 & 18.46 & 26.35 & 34.46 && 17.91 & 4.01 & 15.34 & 35.09 && 22.03 & 6.79 & 22.89 & 34.63 \\
Short Exposures ($N=4$) & 30.81 & 18.41 & 26.53 & 36.02 && 18.46 & 4.33 & 15.97 & 35.86 && 22.17 & 6.87 & 23.91 & 35.11 \\
\midrule
Multi-Exposure Ensemble ($N=2$) & 33.76 & 14.67 & 27.64 & 40.81 && 19.36 & 5.11 & 17.23 & 37.66 &&  23.11 & 8.01 & 25.87 & 36.09\\
Multi-Exposure Ensemble ($N=4$) & \textbf{36.17} & 14.15 & \textbf{29.04} & \textbf{42.17} && \textbf{20.97} & \textbf{5.38} & \textbf{19.46} & \textbf{38.95} && \textbf{24.71} & \textbf{9.13} & \textbf{27.08} & \textbf{37.79} \\
\bottomrule
\end{tabular}
\caption{\textbf{Object Detection Results}: AP results on REDS, MSCOCO, and CityScapes datasets. Our approach of Multi-Exposure Ensemble (Ours) outperforms all baselines.}
\label{tab:detectionresults}
\end{table*}

\medskip
\noindent \textbf{Datasets and Metrics}.
We evaluate our approach using three object detection datasets: Cityscapes~\cite{cordts2016cityscapes}, MS-COCO~\cite{lin2014microsoft} and REDS~\cite{Nah_2019_CVPR_Workshops_REDS}. Cityscapes consists of street scenes captured from a vehicle and consists of 8 categories related to autonomous driving with 2975 training and 500 test images. MS-COCO consists of 80 categories for general object detection with 118k training and 5k validation images. REDS consists of 120fps video sequences of 270 scenes captured by a high speed camera. Dataset represents images with common objects (like person, car, chair etc.).

The ground truth annotations provided in Cityscapes and MS-COCO are used for evaluation. We follow common conventions, train our models on their training sets, and report results on the validation sets. In contrast, REDS does not have object annotations. We thus use a pretrained Faster R-CNN object detector model~\cite{ren2015faster} available in the Detectron2 platform~\cite{wu2019detectron2} to obtain pseudo-ground truth annotations to create our evaluation benchmark containing 270 images with 2160 box annotations.

All results are reported using mean average precision (mAP) across multiple intersection-over-union (IoU) thresholds, following the COCO evaluation protocol~\cite{lin2014microsoft}. 

\medskip
\noindent \textbf{Low-light and Motion Blur Dataset Generation}.
All three datasets mentioned above contain images captured in sufficient light and no noticeable motion blur (scene or camera). 
Since there is no publicly available large-scale annotated dataset containing images captured in low-light and motion blur conditions, we simulate such conditions using various strategies, as described below.

\begin{figure}[htp!]
\centering
\begin{subfigure}{\linewidth}
\includegraphics[width=\linewidth]{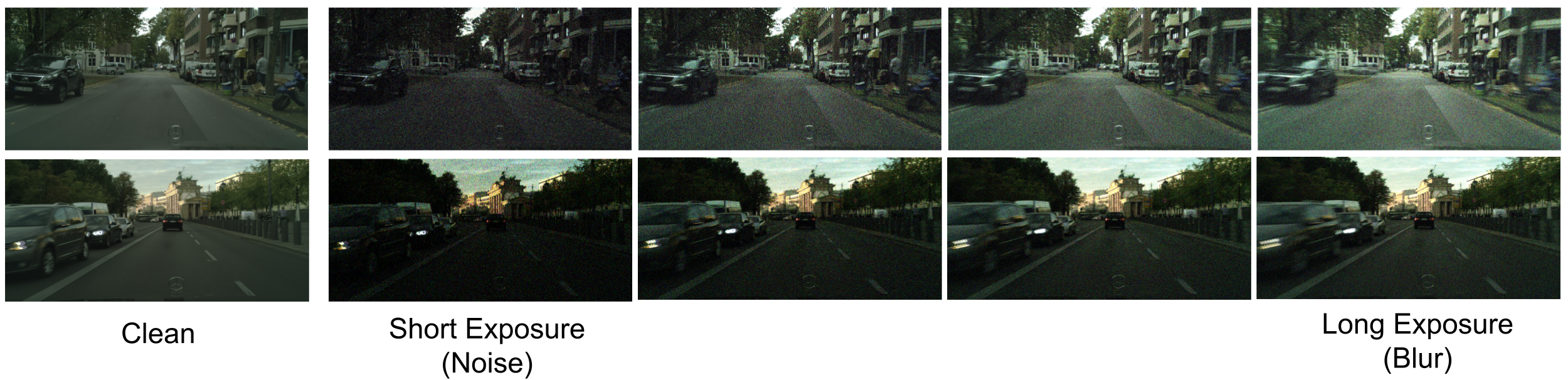}
\vspace{-10pt}
\caption{CityScapes}
\label{fig:cityscapesexample}
\end{subfigure}
\begin{subfigure}{\linewidth}
\vspace{10pt}
\includegraphics[width=\linewidth]{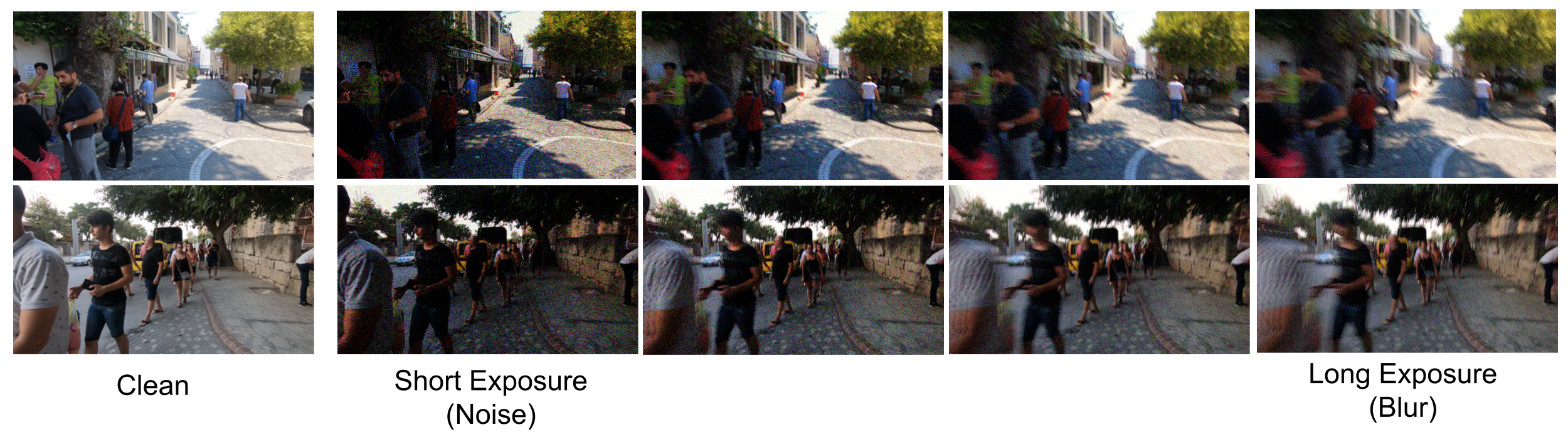}
\vspace{-10pt}
\caption{REDS}
\label{fig:redsexample}
\end{subfigure}
\begin{subfigure}{\linewidth}
\vspace{10pt}
\includegraphics[width=\linewidth]{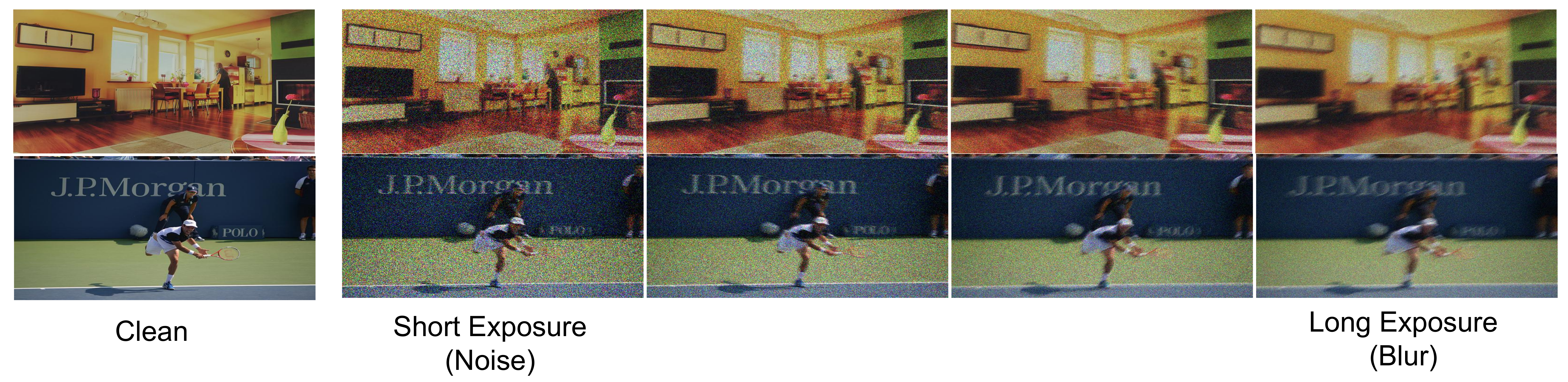}
\vspace{-10pt}
\caption{MS-COCO}
\label{fig:cocoexample}
\end{subfigure}
\begin{subfigure}{\linewidth}
\vspace{10pt}
\includegraphics[width=\linewidth]{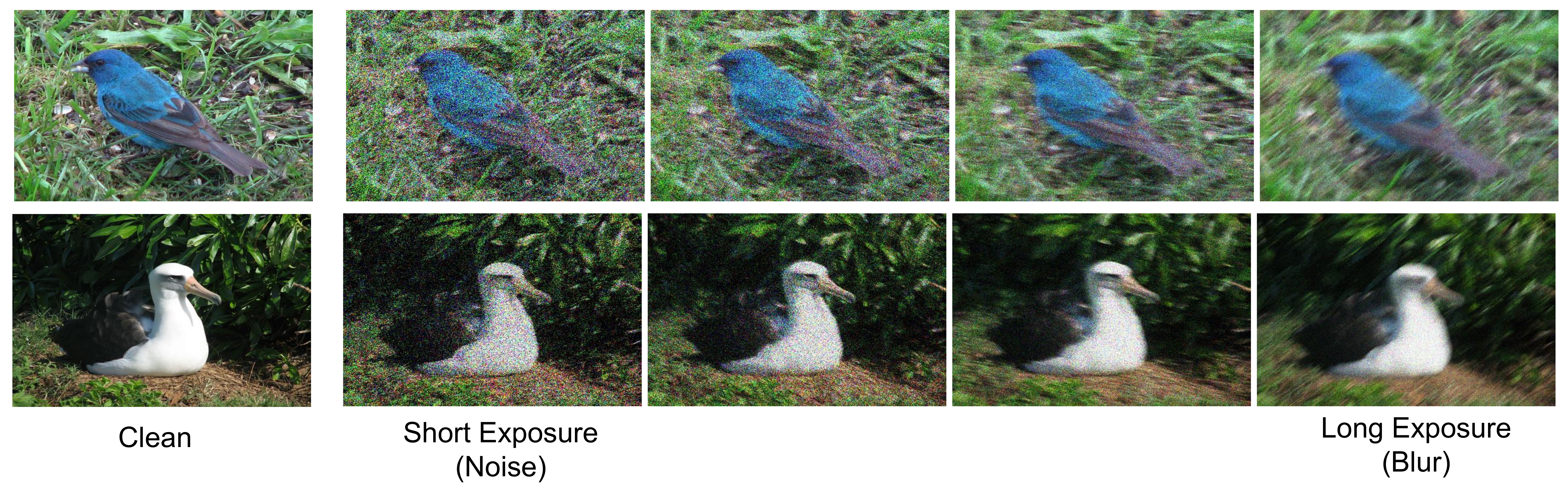}
\vspace{-10pt}
\caption{CUB-200-2011}
\label{fig:birdsexample}
\end{subfigure}
\caption{\textbf{Simulated Images:} Few examples of images with simulated noise and blur. CityScapes and REDS dataset images are generated by simulating low-light frames from high speed video sequence. MS-COCO and Birds dataset images are generated using a single frame by adding noise (shot noise and read noise) and blur (random motion blur kernel) of varying amounts.
}
\label{fig:examples}
\end{figure}

\squishlist
    \item {\it REDS}: Since the REDS dataset contains video sequences captured by a 120fps camera, we first simulate low-light conditions for each individual frame of the sequence by adding Poisson noise (shot noise) and read noise. Multiple frames are then averaged together to generate images with motion blur that captures realistic motion conditions of camera or scene. In practice, we select a random frame from each video sequence, select a varying number of adjacent frames (from 0 to 3 on each side of the frame), and compute their average (after adding noise) to simulate blurry images with motion. This generates images with different exposures, examples of which are shown in Figure~\ref{fig:redsexample}.
    
    \item {\it CityScapes}: CityScapes provides low-fps video sequences around each annotated frame in the dataset (30-frame sequence captured at 17fps). We use a pretrained video interpolation network~\cite{sim2021xvfi} to synthesize high-fps video sequence by increasing the frame rate by a factor of 4x. A motion-blurred image is then generated as with the REDS dataset, that is adding noise to each individual frame, and averaging a number of adjacent frames. Figure~\ref{fig:cityscapesexample} shows examples of simulated low-light and motion-blur frames used for training and evaluation on the CityScapes dataset. The resulting images indeed represent realistic motion conditions under autonomous driving scenarios (like fast moving camera/car or moving pedestrians, other vehicles etc.).

    \item {\it MS-COCO}: As the MS-COCO dataset does not contain any video sequences, we simulate the blur and noise from a single image using the same procedure as the image corruptions benchmark in~\cite{hendrycks2018benchmarking} by selecting varying severity of shot noise and motion blur. Specifically, the noisiest image has a shot noise level of 4 and a motion blur level 1. Subsequent levels in the dual corruptions are simulated by increasing the motion blur and decreasing the shot noise successively to generate 4 levels of dual corruptions. Figure~\ref{fig:cocoexample} shows a few examples of simulated images. We note that, contrary to the other two datasets above, the blur simulated by this approach is not spatially varying.
\squishend

\medskip
\noindent \textbf{Baselines}.
We compare our approach with the following set of baselines. All approaches use the same backbone for fair comparison. We evaluate all the methods using all four exposures and report the results for the best exposure settings.
\squishlist
    \item {\it Clean Model}: This baseline model is trained only on clean images, and evaluated on noisy images.
    \item {\it Stylized Training}: We follow the data augmentation approach of~\cite{michaelis2019benchmarking}, who propose to augment training images with stylization for robustness.
    \item {\it Single Exposure}: We train a model on a dataset containing varying exposures and clean images, essentially considering distortions as a way to perform data augmentation~\cite{hendrycks2018benchmarking}. For evaluation, we select the single exposure setting yielding the best performance and report those results. This baseline acts as an oracle for the selecting the best performing exposure time at inference time.
    \item {\it Denoising}: This baseline represents the conventional approach of denoising the noisy images under low-light conditions. We perform both training and inference on denoised images. Here, we experiment with the BM3D~\cite{dabov2007image} and MPRNet~\cite{Zamir2021MPRNet} approaches for denoising the images. 
    \item {\it Deblurring}: We also compare with the approach of debluring the images for scene inference, where we use a deblurring model~\cite{carbajal2021non}. We perform both training and evaluation of our model using deblurred images.
    \item{\it Denoising + Deblurring}: As the test images in low-light and motion blur have both noise and blur, we also compare with the approach of denoising (BM3D) followed by deblurring. Model is trained and evaluated using Denoised+Deblurred images.
    \item{\it Short Exposures}: This baseline compares with the approach of evaluating using multiple short exposures by using the ensemble prediction from $N$ short exposure images. Model is trained with short exposure images.
\squishend

\medskip
\noindent \textbf{Implementation Details}.
We used the official implementation of the FCOS architecture~\cite{tian2019adelaidet} for the object detection experiments, which is based on the Detectron2 framework~\cite{wu2019detectron2}. ResNet-50~\cite{he2016deep} with FPN was used as backbone for training and initialized with ImageNet pretraining weights for all our models.
We followed the hyperparameters from Detectron2 to train our models. MS-COCO models were trained with a learning rate of 0.01, batch size of 16 for 90k iterations whereas CityScapes model were trained with a learning rate of 0.005, batch size of 8 for 24k iterations. REDS is used only for evaluation, in this case we use the model trained on MS-COCO.

\begin{figure}
\begin{center}
\includegraphics[width=\linewidth]{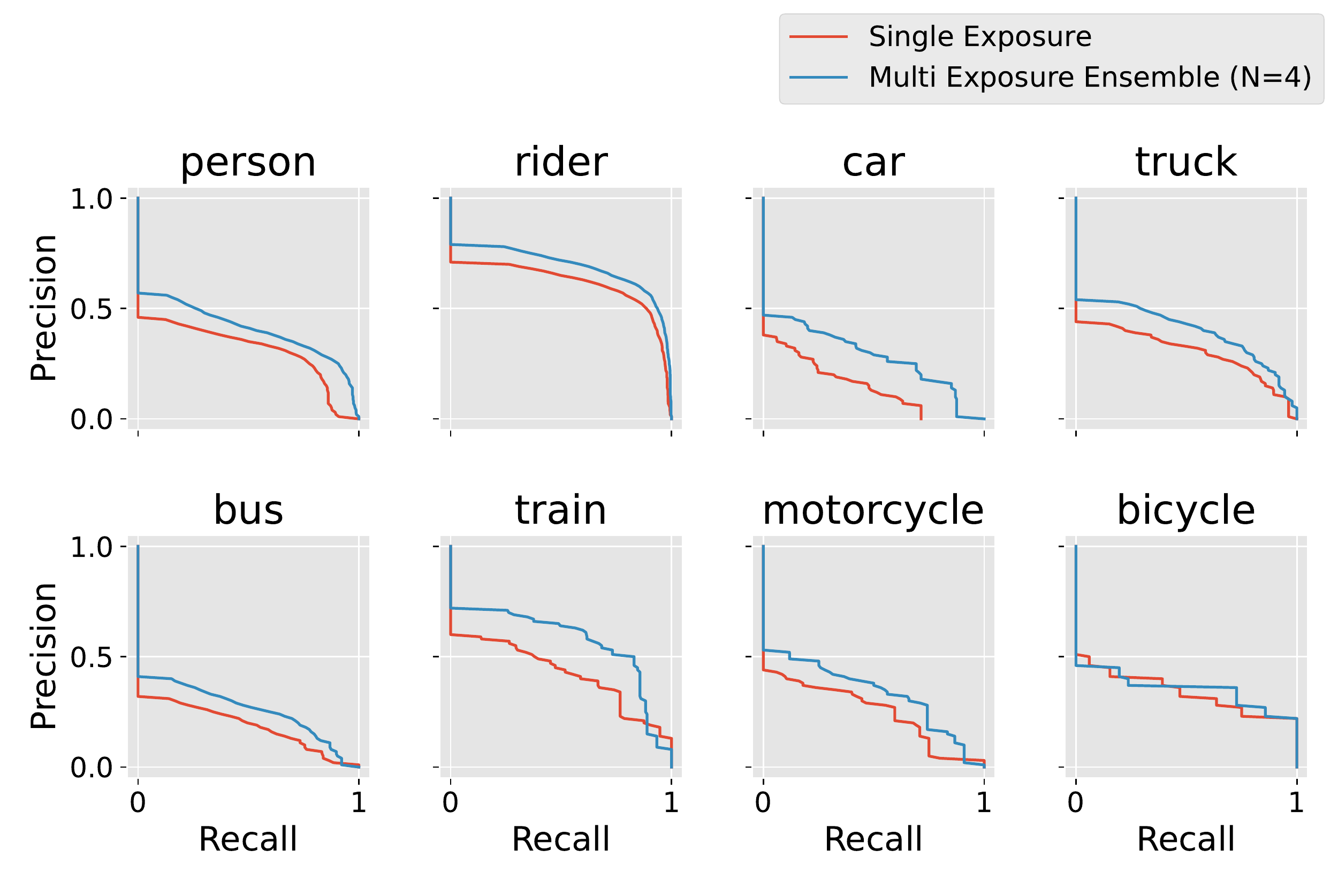}
\end{center}
\caption{\textbf{Precision Recall Curve} of our approach and baselines on CityScapes Dataset for all 8 categories with IOU threshold of 0.5. We see significant improvement for `person' and `car' categories, which are most common in the dataset.}
\label{fig:prcurve}
\end{figure}

\begin{figure*}
\centering
\begin{subfigure}{0.97\linewidth}
\includegraphics[width=\linewidth]{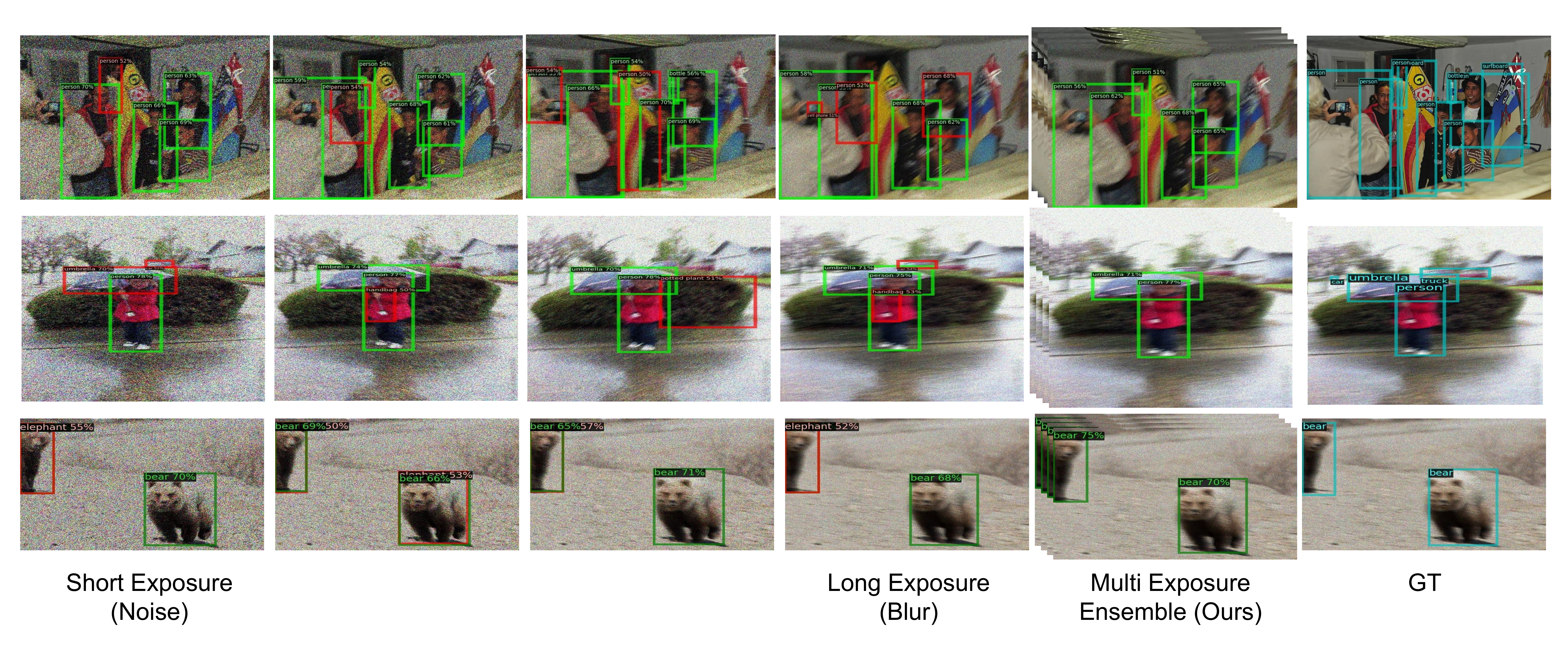}
\vspace{-20pt}
\caption{MS-COCO Dataset Results}
\end{subfigure}
\begin{subfigure}{0.96\linewidth}
    \includegraphics[width=\linewidth]{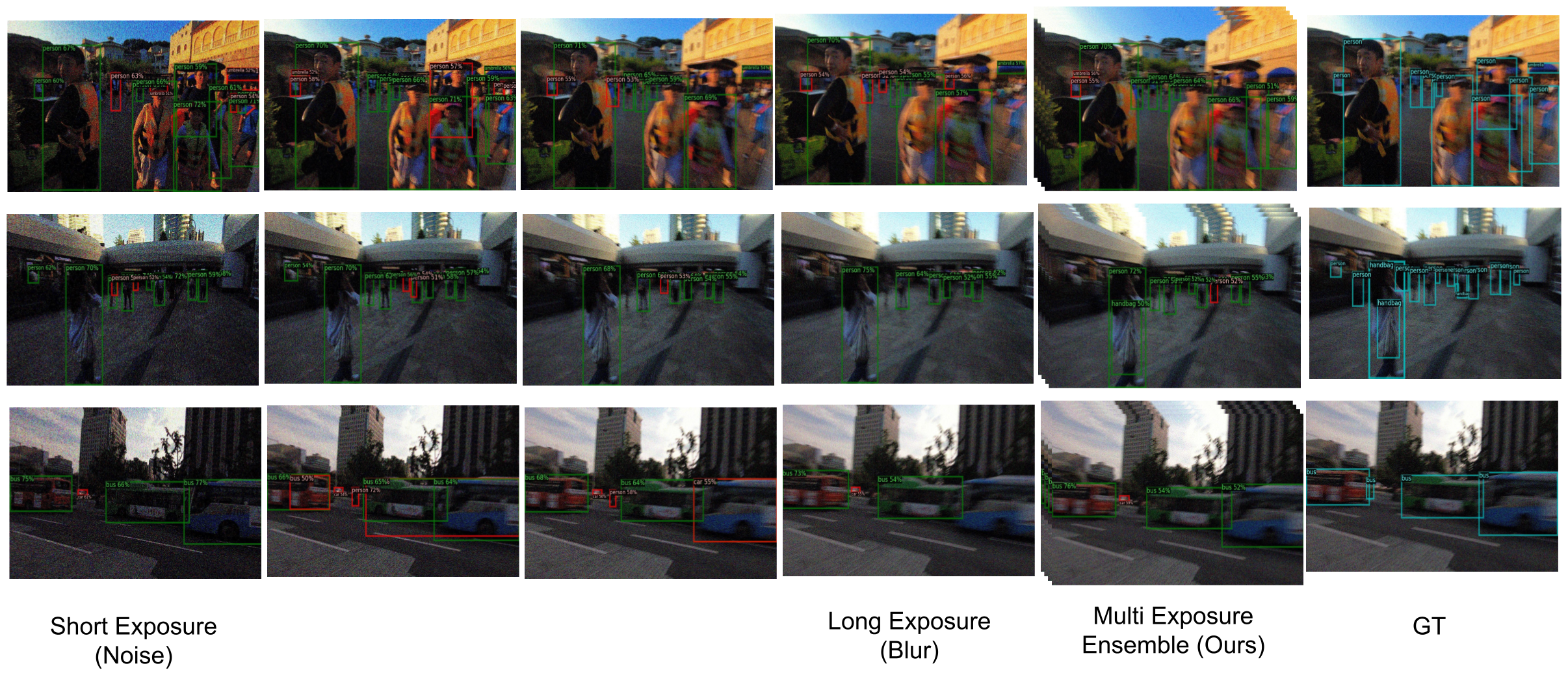}
\vspace{-15pt}
\caption{REDS Dataset Results}
\end{subfigure}
\begin{subfigure}{0.97\linewidth}
\includegraphics[width=\linewidth]{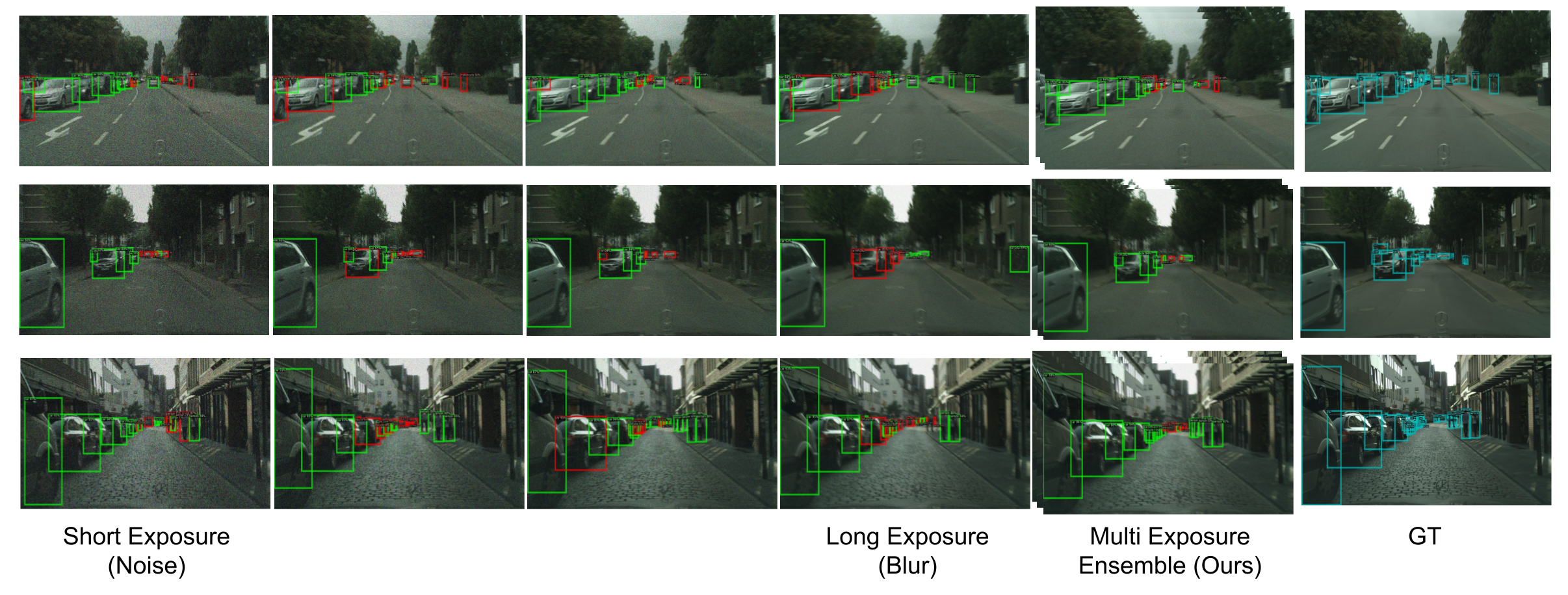}
\vspace{-15pt}
\caption{CityScapes Dataset Results}
\end{subfigure}
\caption{\textbf{Object Detection Results} for MS-COCO, REDS and CityScapes Dataset. Correct/Incorrect predictions are highlighted with green/red and ground truth boxes are highlighted with blue in the clean image. First 4 columns show results on single captures followed by a column with results from multi-exposure captures using our approach. Single Captures have a lot more false positives (red) while our approach effectively removes those cases (Better viewed on screen).
}
\label{fig:detectionresults}
\end{figure*}

\medskip
\noindent \textbf{Results and Discussions}.
Table~\ref{tab:detectionresults} shows the results (in mAP along with AP of small, medium and large objects) of our approach on all three datasets. Our method outperforms all baselines by a significant margin. Our approach beats Single Exposure baseline by 6\% in REDS, 2.9\% in CityScapes, and 3.5\% in MS-COCO with four exposures. In other words, it is best to leverage all the dual-corruption images even if we knew the best possible single exposure ahead of time.
Denoising provides improvements over Single Exposure baseline in some cases but is not as effective.
Deblurring approaches does not show performance improvement over Single Exposure baseline in most cases. This is because images contain both noise and blur and deblurring models are specialized to handle only blur. Deblurring+Denoising baseline also shows relatively minor performance gain. We see significant gain with Short Exposures (with 4 exposures) baseline, highlighting the benefit of ensemble prediction. However, since all the exposures are short, they all suffer from sever noise and have similar errors, and hence outperformed by our method. Our method provides large improvements even with two exposures, and increasing the number of exposures (from two to four) further increases the performance. This highlights that our approach benefits with more number of exposures as different exposures have a wide variety of dual corruption level.

Figure~\ref{fig:detectionresults} shows representative qualitative examples of our approach for object detection and shows direct comparison with each individual exposure and its predictions.
The correct/incorrect bounding boxes are highlighted in green/red and ground truth bounding boxes are highlighted in blue on the clean image (right). 
Our approach makes fewer false positive predictions (red) compared to the Single Exposure. Since individual single captures make different false positive predictions, the ensemble is able to remove those false positive boxes. 
Figure~\ref{fig:prcurve} shows the precision recall curve on CityScapes dataset for IOU threshold of 0.5 for all 8 categories in the dataset. We see a significant improvement in area under the curve for person and car category, which is the most common in the dataset.

\begin{table}
\centering
\begin{tabular}{l c c}
\toprule
Method & Top-1 & Top-5\\
\midrule
Clean Model & 6.13 & 13.45 \\
Stylized Training~\cite{michaelis2019benchmarking} & 9.51 & 17.83 \\
Single Exposure & 41.18 & 64.13\\
Denoising (BM3D)~\cite{dabov2007image} & 43.34 & 67.11\\
Deblurring~\cite{carbajal2021non} & 39.13 & 60.45 \\
Denoising~\cite{dabov2007image} + Deblurring~\cite{carbajal2021non} & 42.95 & 66.59 \\
Short Exposures ($N=4$) & 45.16 & 69.84 \\
\midrule
Multi Exposure Ensemble ($N=2$) & 52.10 & 74.13 \\
Multi Exposure Ensemble ($N=4$) & \textbf{55.27} & \textbf{79.34} \\
\bottomrule
\end{tabular}
\caption{\textbf{Image Classification Results}: Top-1 and top-5 accuracy results on CUB-200-2011 dataset. Our approach of Multi-Exposure Ensemble outperforms all the baselines.}
\label{tab:clsresults}
\end{table}

\subsection{Image Classification} 
\label{sec:classification}

\noindent \textbf{Instantiation}. Similar to object detection, our approach uses a shared CNN architecture as a feature extractor. In particular, we used a ResNet-18~\cite{he2016deep} as the image classification architecture. The feature consistency loss $\ell_\mathrm{fc}$ is defined as the L2 distance between the feature map output of the final layer (after global average pooling) to encourage consistent predictions. The model returns the average of the predictions (probability output) from multiple degraded images (as the ensemble operator $\mathcal{G}$) for the final output.

\medskip
\noindent \textbf{Datasets, Metrics, and Baselines}.
\label{sec:simulateddataset}
We use simulated images from the CUB-200-2011 image classification dataset~\cite{WahCUB_200_2011}. CUB-200-2011 is commonly used for fine-grained image classification benchmarks and consists of 200 species of birds with 5,994 training images and 5,794 test images. All results are reported using top-1/5 accuracy on the test set, following the standard evaluation protocol for image classification. A set of baselines similar to the ones used in the experiments on object detection (Section~\ref{sec:detection}) is considered here.

\medskip
\noindent \textbf{Simulating Noise and Blur}. Since CUB only contains single images, we employ the same strategy to generate dual corruption images as for the MS-COCO dataset in the object detection experiments (Section~\ref{sec:detection}). Figure~\ref{fig:birdsexample} shows a few examples of simulated images.

\medskip
\noindent \textbf{Implementation Details}. The model is trained with SGD with momentum of 0.9, base learning rate of 0.1 with cosine decay and batch size of 32 is used to train for 100 epochs.

\medskip
\noindent \textbf{Results and Discussions}.
Table~\ref{tab:clsresults} shows top-1 and top-5 accuracy of our approach on the simulated CUB dataset. We report results of our model using two and four exposure settings. 
Our method outperforms both baselines using a single exposure by a significant margin. Compared to choosing the single best exposure, our approach, with $N=4$, attains an overall gain of 14.1\% and 15.2\% in top-1 and top-5 accuracy respectively.
Our approach shows significant gains with only two exposures however having more number of exposures (from 2 to 4) further helps the overall performance.

\medskip
\noindent \textbf{Ablation Studies}.
\begin{figure}
\centering
\includegraphics[width=0.8\linewidth]{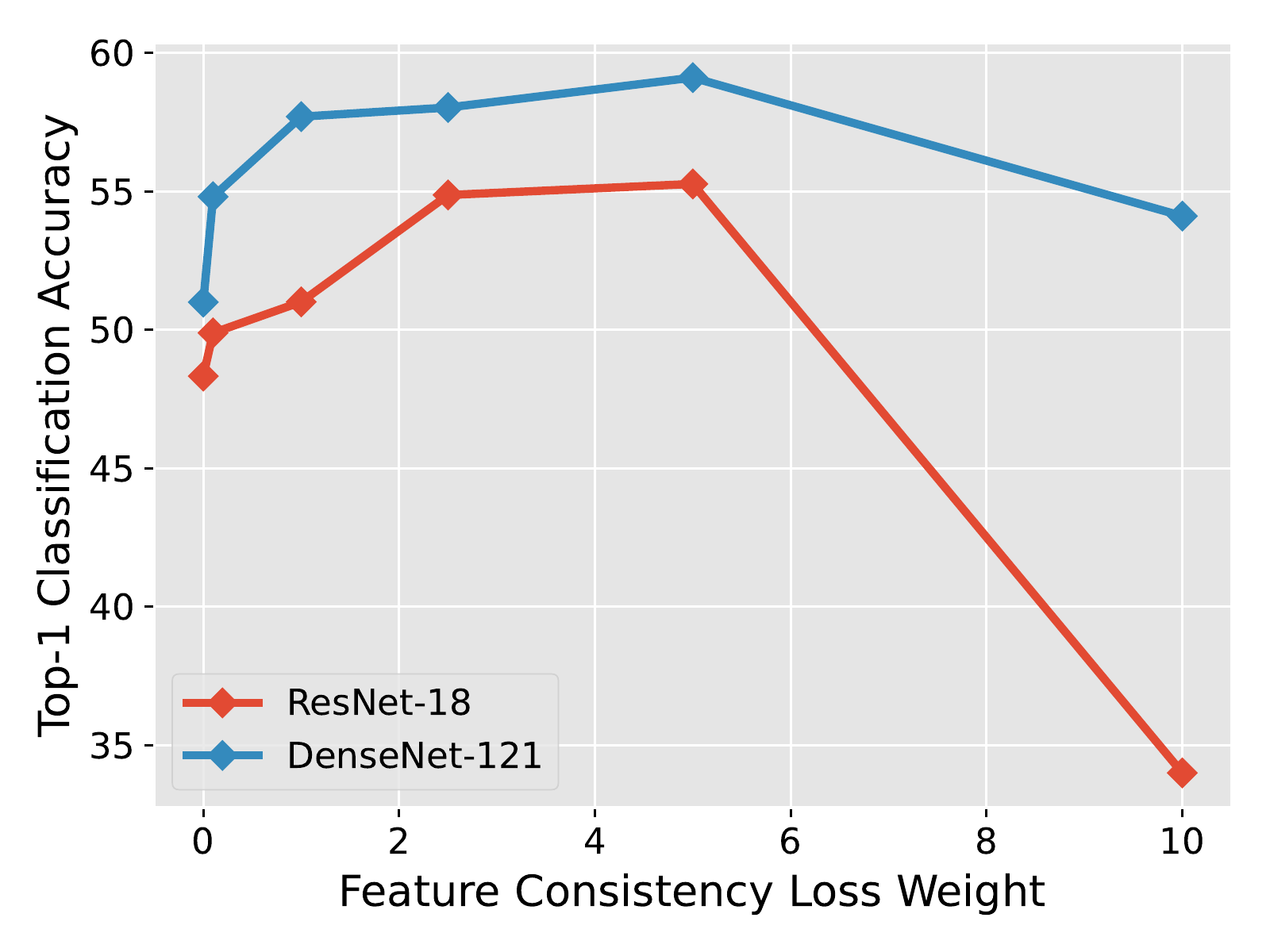}
\caption{\textbf{Ablation Studies:} Image classification results of our approach on CUB-200-2011 while varying feature consistency loss weight and backbone architecture.}
\label{fig:ablation_fcloss}
\end{figure}
We study the performance of our approach with varying weight for feature consistency loss. Figure~\ref{fig:ablation_fcloss} shows that our approach performs best for the weight factor of 5 image classification on CUB-200-2011 Dataset. We also evaluate the performance of our approach with another backbone architecture. Figure~\ref{fig:ablation_fcloss} shows similar performance gain using DenseNet-121~\cite{huang2017densely} which highlights the versatility of our approach as it can extend to different CNN feature extractors.
 
\begin{figure}
\centering
\begin{subfigure}{0.55\linewidth}
    \includegraphics[width=\linewidth]{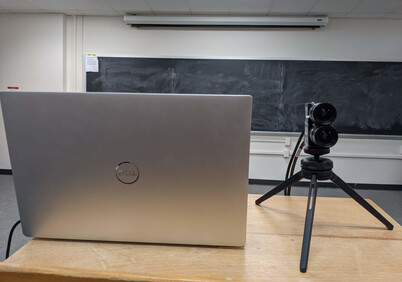}
\end{subfigure}
\begin{subfigure}{0.4\linewidth}
    \includegraphics[width=\linewidth]{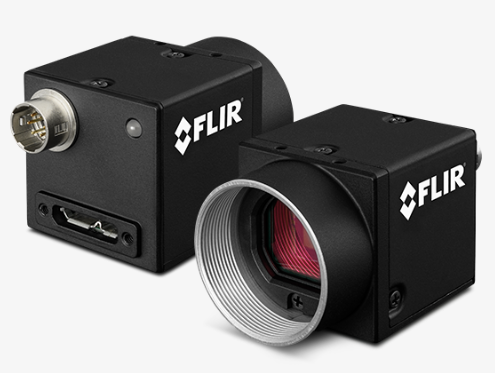}
\end{subfigure}
\caption{\textbf{Camera Setup} for capturing multiple exposure images}
\label{fig:camerasetup}
\end{figure}

\begin{figure}
\centering
\includegraphics[width=\linewidth]{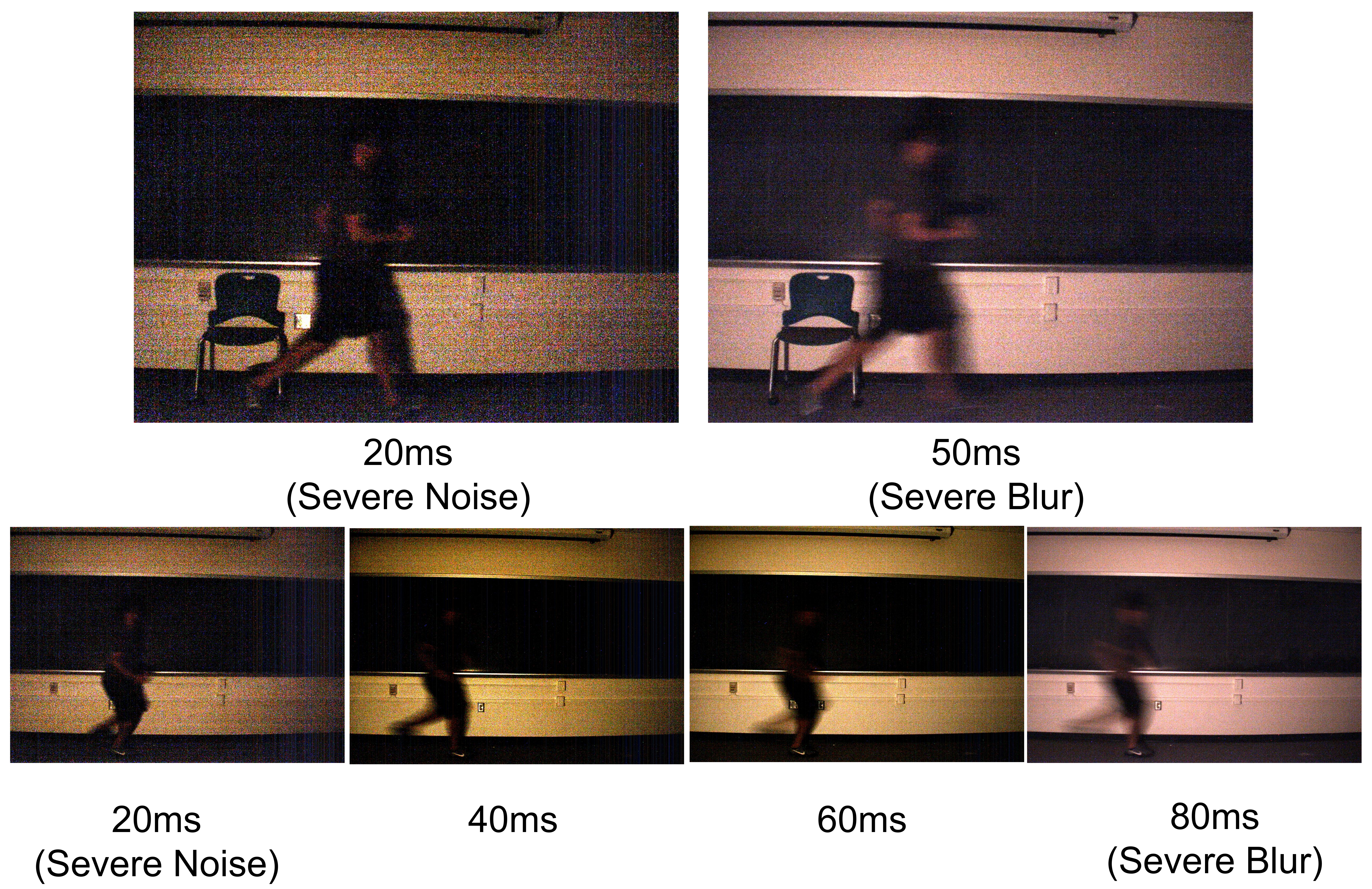}
\caption[]{\textbf{Examples of Real Captures:} Images captured with varying exposure settings with our multi camera setup. Images with shorter exposure have severe noise while images with longer exposure contain motion blur for the moving objects.}
\label{fig:realexamplesupp2}
\end{figure}

\begin{figure*}
\begin{subfigure}{\linewidth}
    \centering
    \includegraphics[width=\linewidth]{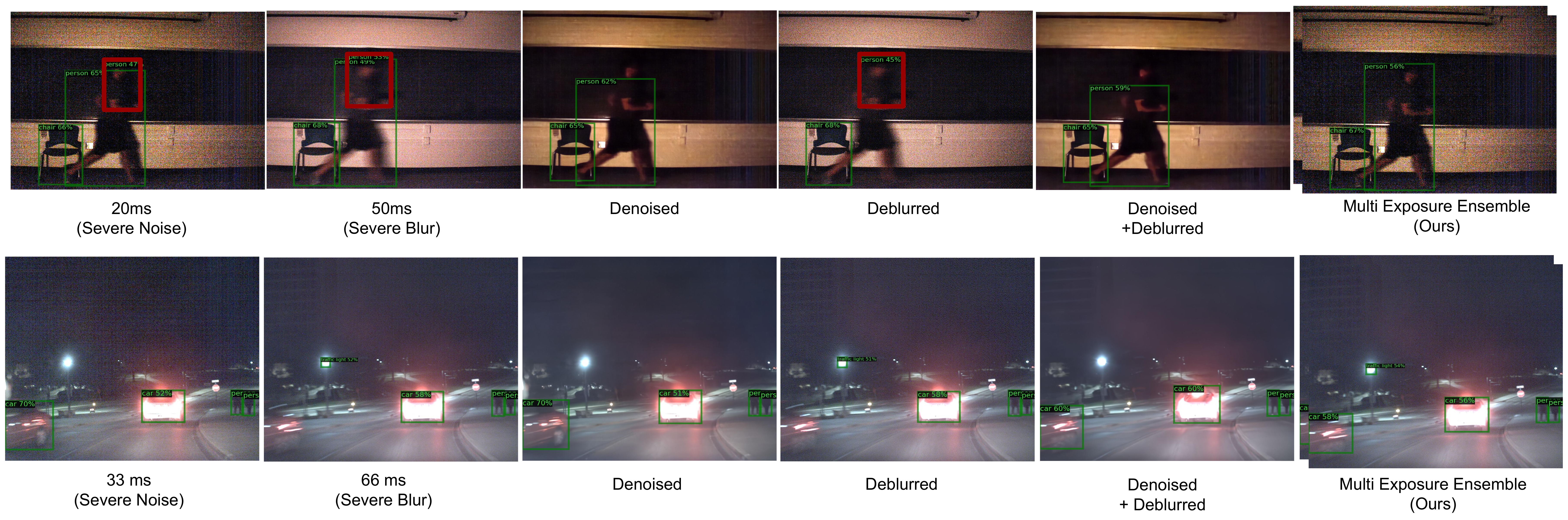}
\end{subfigure}
\caption{\textbf{Object Detection Results on Real Captures:} Scene in the first row contains an indoor scenario with two objects: a person (moving) and a chair (stationary). Single Exposures are severely affected by noise and/or blur: detects false positives or inaccurate bounding boxes. Scene in the second row contains a driving scenario with a car (moving) on the left  and traffic light (stationary) in the front. Single Exposures fail to detect the moving car or the stationary traffic light. Multi-Exposure Ensemble approach (right) leverages multiple exposures and detects all objects with correct labels and tight bounding boxes in both scenes.}
\label{fig:realresults}
\end{figure*}

\section{Experiments with Real Captures}

Finally, we evaluate our approach on real images by capturing multiple simultaneous exposures of the same scene.\medskip

\noindent \textbf{Camera Setup:} Our setup includes four BlackflyS USB3 cameras~\cite{teledyne} by Teledyne Flir. These are machine vision cameras that can capture colored images with a resolution of $1280 \times 1024$ with up to 175 frames per second. Same lenses (Tamron 8mm) are used for all cameras, which are stacked together to get similar (overlapping) fields-of-view. Aside from an approximate physical alignment of the cameras, no further alignment of the captured images is done as all cameras have similar fields-of-view, and the scene is sufficiently far away. Cameras are connected to a computer that triggers the simultaneous captures (software sync). Our complete setup is shown in Figure~\ref{fig:camerasetup}.

We use spinnaker SDK~\cite{spinnaker} provided by Teledyne to capture RAW images. Maximum available gain (18dB) for the camera is used and a gamma correction ($\gamma=2.2$) is applied on the captures to get the final images. We set different exposure times for each camera and synchronously capture images using all the cameras.\medskip

\noindent \textbf{Exposure Selection: } We manually select the exposure times in order to span a wide range of exposures while ensuring that images are not too under- nor over-exposed. Our indoor scenes consist of fast moving objects in a very dark environment ($\sim$0.25lux) lit by a single light source. 
We experiment with multiple settings depending on the lightning conditions including A) 20-30-40-50ms, B) 20-40-60-80ms, and C) 16-33-66-100ms. When evaluating our approach, we use two or four exposures, examples of which are shown in Figure~\ref{fig:realexamplesupp2}.\medskip

\noindent \textbf{Results and Discussions}.
We train our object detection models with the simulated images from MS-COCO dataset and evaluate the trained model on real captures. 
Figure ~\ref{fig:realresults} shows sample results with the real captures on two scenes. Both scenes consists of both fast moving and stationary objects under low-light. The prediction output from the individual exposure contain several false positives and inaccurate boxes. By leveraging the multiple exposures across the space of dual corruptions, our method is able to correctly detect all the objects with tight bounding boxes and remove false positive boxes.

Our approach shows performs better inference even with two exposures ($N=2$). As we increase the number of exposures, the prediction improves as long as the exposures are not too noisy or blurry for inference (as that can deteriorate the performance of the ensemble prediction).
One simple heuristic that performs well with our approach is to select exposure times around the \emph{auto-exposure} value, as this ensures the frames are not too under- or over-exposed.
We show more examples in the supplementary text with two and four exposures including failure cases.

 \section{Discussion and Future Outlook}
\noindent {\bf Multi-Exposure Cameras:} We demonstrated our approach of multi exposure captures by utilizing multiple cameras with similar (overlapping) fields-of-view. With cameras that are capable of capturing multiple images with varying exposures simultaneously~\cite{Nayar:2000, nguyen2022learning}, multiple exposure images could be captured with a single camera, thus making it easier to perform spatio-temporal alignment. Our work can be considered as a preliminary proof-of-concept for an eventual implementation where a single camera can capture multiple exposure images. Demonstrating and evaluating our approach on such images is an important next step.
\medskip

\noindent {\bf Exposure Selection for Multiple Captures:}
Most modern cameras have the functionality of \emph{auto-exposure} that selects the exposure setting based on the lighting and motion conditions (light and motion metering) of the scene for the best image quality.
The optimal exposure for inference is a function of the amount of light and motion (camera/scene) in the scene, and determining it automatically (for a single exposure) is an active area of research~\cite{Onzon2021}.
With the ability to capture multiple exposures, an important research problem is to develop generalized auto-exposure techniques for \emph{multiple captures} that result in the best performance for the inference tasks under these challenging conditions.
\medskip

\noindent {\bf Computational Considerations:} 
Capturing, processing and performing inference on multiple exposures incurs a linear increase in computational cost. However, since many of these computations can be done in parallel, the increase in latency is small which is important for safety critical applications like autonomous driving. Our approach is agnostic to the number of exposures during inference, which allows inference systems to switch between multi-exposure settings (during challenging conditions of low-light and/or motion) and single exposure setting during less challenging conditions (day-time driving or slow/no motion). In practice, the inference system can operate at no computational overhead by using single exposure setting during most of the time (\eg daytime driving) and use multi-exposure setting during more challenging conditions (\eg night time driving).
\medskip

\noindent {\bf Dual Image Degradations:} So far, we have considered the dual corruptions of noise and blur. In principle, a similar dual relationship exists between several other image degradation pairs, such as, rain and defocus blur~\cite{Garg:2005}, and snow and motion blur~\cite{Barnum:2008}. A promising research direction is to evaluate the proposed approach on other such dual pairs of image degradations, toward the goal of achieving `all-weather' computer vision systems.

\ifpeerreview \else
\section*{Acknowledgments}
This research was supported in part by the National Science Foundation under the grants CAREER \#1943149 and \#2003129, Intel MLWiNS grant, and a SONY Faculty Innovation Award.
\fi

\bibliographystyle{IEEEtran}
\bibliography{references}

\ifpeerreview \else

\vfill
\begin{IEEEbiographynophoto}{Bhavya Goyal} 
is a PhD student in CS at University of Wisconsin-Madison, advised by Prof. Mohit Gupta. He received his bachelors in CS from Indian Institute of Technology, Delhi in 2016. His interests broadly include computer vision and computational imaging. He is particularly interested in learning based approaches for emerging sensing technologies.
\end{IEEEbiographynophoto}

\begin{IEEEbiographynophoto}{Jean-François Lalonde} is an Associate Professor in the Electrical and Computer Engineering Department at Université Laval. Previously, he was a Post-Doctoral Associate at Disney Research, Pittsburgh. He received a Ph.D. in Robotics from Carnegie Mellon University in 2011. His research interests lie at the intersection of computer vision, computer graphics, and machine learning. In particular, he is interested in exploring how physics-based models and data-driven machine learning techniques can be unified to better understand, model, interpret, and recreate the richness of our visual world. To this end, he has published 70 refereed papers which have been cited close to 4,000 times. He is actively involved in bringing research ideas to commercial products, as demonstrated by his 9 patents, several technology transfers with large companies such as Adobe and Meta, and involvement as scientific advisor for several high tech startups.
\end{IEEEbiographynophoto}

\begin{IEEEbiographynophoto}{Yin Li} is an Assistant Professor in the Department of Biostatistics and Medical Informatics and affiliate faculty in the Department of Computer Sciences at the University of Wisconsin-Madison. Previously, he obtained my PhD from the Georgia Institute of Technology and was a Postdoctoral Fellow at the Carnegie Mellon University. His primary research focus is computer vision. He is also interested in the applications of vision and learning for mobile health. Specifically, his group develops methods and systems to automatically analyze human activities for healthcare applications. \end{IEEEbiographynophoto}

\begin{IEEEbiographynophoto}{Mohit Gupta} is an assistant professor of Computer Science at the University of Wisconsin-Madison. Before coming to Madison, he was a research scientist in Columbia University. He received his PhD from the Robotics Institute, Carnegie Mellon University. He directs the WISIONLab with research interests broadly in computer vision and computational imaging.
\end{IEEEbiographynophoto}
\vfill

\onecolumn
\section{Supplementary Report: Robust Scene Inference under Noise-Blur Dual Corruptions}
This document provides additional results that are not included in the main paper. 

\subsection{Results for Object Detection}
\noindent \textbf{Comparison to Baselines}: We compare our approach with additional baselines. Table~\ref{tab:suppdetectionresults} shows performance of the model trained and evaluated on clean images.
We also show the results of training and testing with a single corruption level. Results are included for four different noise-blur dual corruption levels (from 1 to 4) with increasing motion blur and decreasing the shot noise image.
Comparing with clean images shows the impact of noise and blur degradation as the mAP drops significantly.
Our approach utilizes clean images and corrupted images with feature consistency that helps the model to learn robust features. Our model outperforms these baselines by a significant margin using the same model capacity.\smallskip

\begin{table*}[htp]
\centering
\begin{tabular}{l | c c c c c | c c c c c | c c c c}
\toprule
\multirow{2}{*}{Method} & \multicolumn{4}{c}{REDS} && \multicolumn{4}{c}{{CityScapes}} &&  \multicolumn{4}{c}{{MS-COCO}}\\
\cline{2-6}
\cline{7-11}
\cline{12-15}
 & mAP & APs & APm & APl && mAP & APs & APm & APl && mAP & APs & APm & APl\\
\midrule
Clean Training \& Testing & 78.21 &52.94 &73.91&84.33 && 33.36 & 10.40 & 32.26 & 54.70 && 38.59 & 22.9 & 42.28 & 49.56\\
\midrule
Corruption Level 1 (Severe Noise) & 23.46 & 16.14 & 24.08 & 26.27 && 14.06 & 1.82 & 11.80 & 30.98 && 20.26 & 6.18 & 21.18 & 32.73\\
Corruption Level 2 & 30.20 & 20.27 & 25.75 & 36.88 && 17.19 & 3.71 & 15.36 & 33.84 &&  20.29 & 5.59 & 21.19 & 32.21\\
Corruption Level 3 & 27.85 & 19.75 & 23.80 & 35.09 && 17.07 & 3.26 & 15.45 & 32.89 &&  20.21 & 6.35 & 20.94 & 32.70\\
Corruption Level 4 (Severe Blur) & 26.78 & 15.51 & 20.13 & 33.53 && 15.94 & 4.21 & 14.73 &30.39 && 20.47 & 6.45 & 20.94 & 32.32\\
\midrule
Multi-Exposure Ensemble ($N=4$) & \textbf{36.17} & 14.15 & \textbf{29.04} & \textbf{42.17} && \textbf{20.97} & \textbf{5.38} & \textbf{19.46} & \textbf{38.95} && \textbf{24.71} & \textbf{9.13} & \textbf{27.08} & \textbf{37.79} \\
\bottomrule
\end{tabular}
\caption{\textbf{Object Detection Results}: AP results on REDS, MSCOCO, and CityScapes datasets.}
\label{tab:suppdetectionresults}
\end{table*}

\noindent \textbf{Results Visualization for Object Detection:} Figure~\ref{fig:cocoresultssupp} shows examples where our approach outperforms the baselines. The first row of Figure~\ref{fig:cocoresultssupp} shows an example where one baseline predicts correct bounding boxes and our approach is as good as best single exposure baseline. Figure~\ref{fig:realresultssupp} shows a few result images with real captures using our approach and the baseline. Our method is more effective in predicting the correct bounding boxes with fewer false positive boxes.
\smallskip

\noindent \textbf{Failure Cases for Object Detection:} Figure~\ref{fig:cocofailuresupp} and \ref{fig:realfailuresupp} shows some failure cases where our approach performs worse than single exposure baseline. Since our approach relies on the average of output predictions, it fails to perform well when one of the exposure has too much degradation.
\smallskip

\begin{figure*}[htp]
\centering
\includegraphics[width=.9\linewidth]{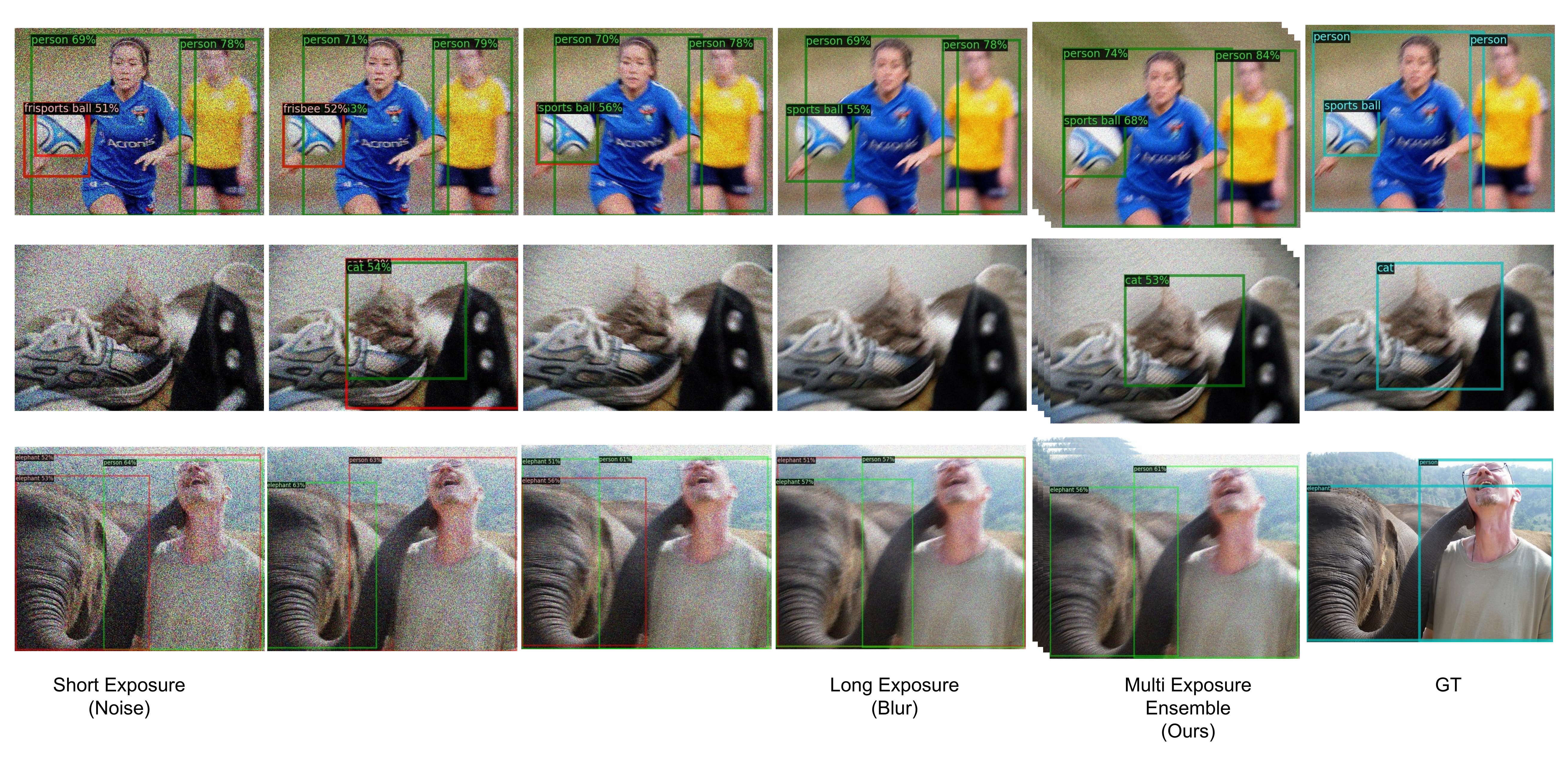}
\caption[]{\textbf{Object Detection Results on MS-COCO dataset:} Correct/Incorrect predictions are highlighted with green/red and ground truth boxes are highlighted with blue in the clean image. Single Exposures have a lot more false positives (red) while our approach effectively removes those cases. For the first scene, our approach produces tighter bounding boxes than individual predictions (Better viewed on screen).
}
\label{fig:cocoresultssupp}
\end{figure*}

\begin{figure*}[htp]
\centering
\includegraphics[width=.9\linewidth]{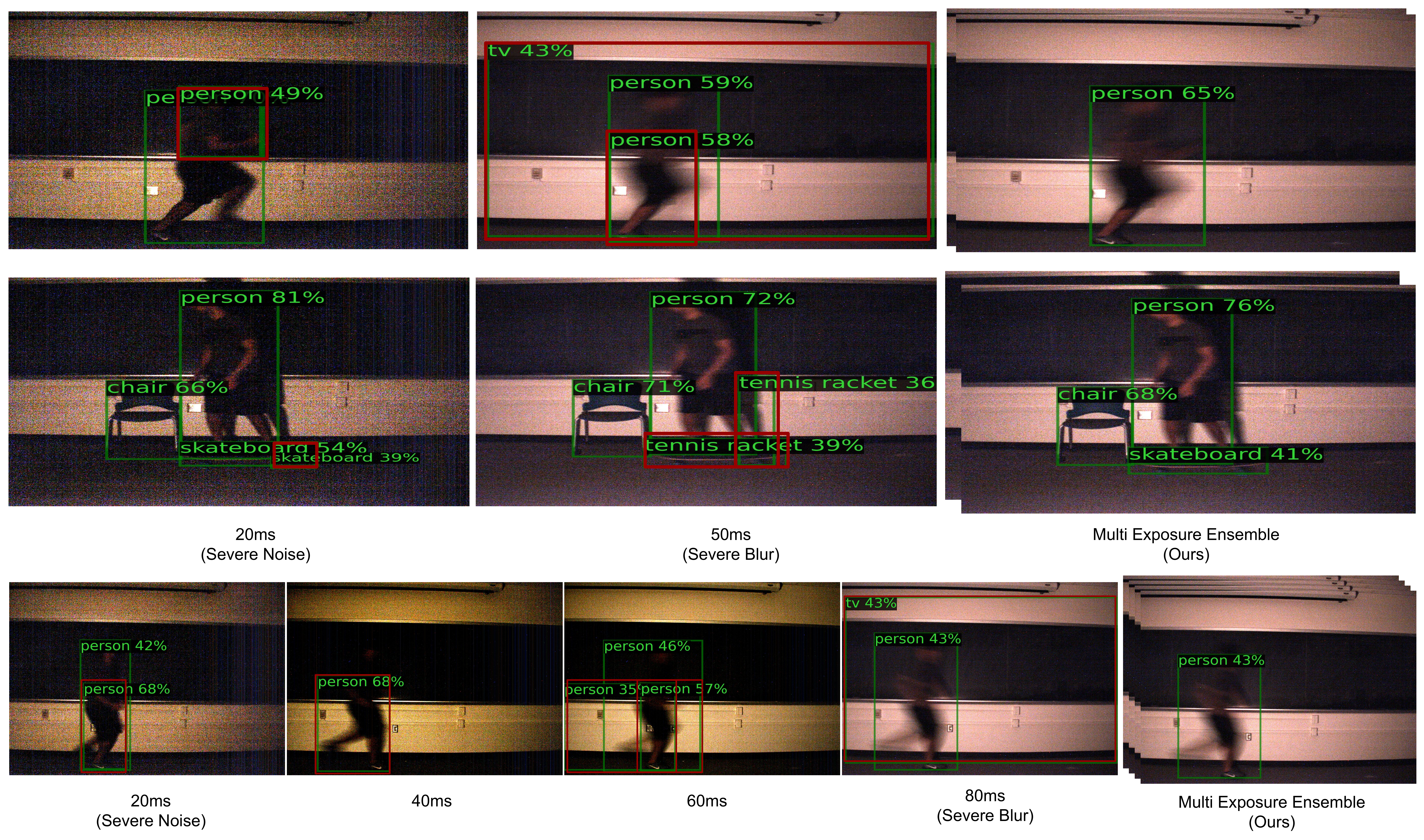}
\caption[]{\textbf{Object detection results with Real Captures:} Single Exposures are severely affected by noise and/or blur. The model detects false positives and inaccurate bounding boxes. Multi-Exposure Ensemble approach (right) leverages multiple exposures and detects all objects with correct labels and tight bounding boxes}
\label{fig:realresultssupp}
\end{figure*}

\begin{figure*}[htp]
\centering
\includegraphics[width=.9\linewidth]{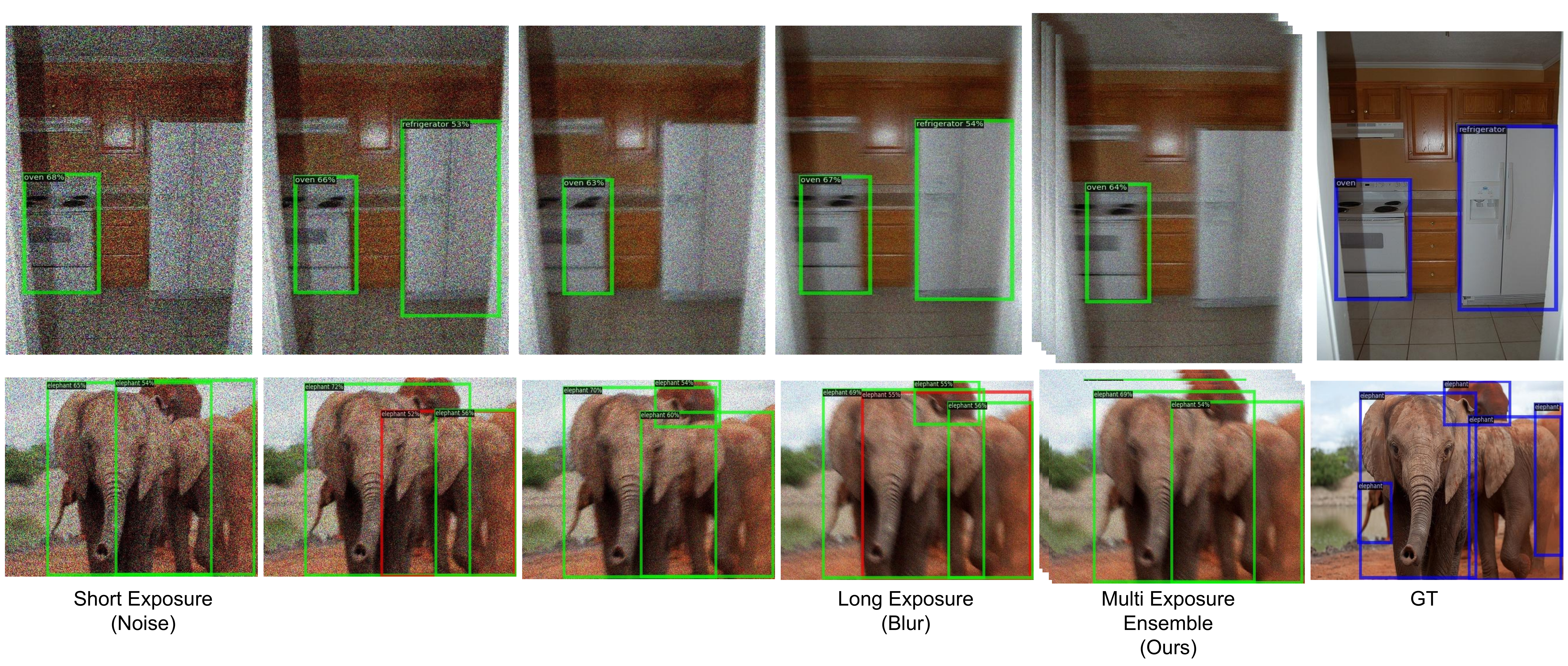}
\caption[]{\textbf{Object Detection Failure Cases on MS-COCO dataset:} Figure shows examples where single exposure performs better than our approach. First scene contains two objects and our approach fails to detect second object. Second scene contains a lot of overlapping ground truth bounding boxes and our approach fails to detects a few bounding boxes.}
\label{fig:cocofailuresupp}
\end{figure*} 

\begin{figure*}[htp]
\centering
\includegraphics[width=.9\linewidth]{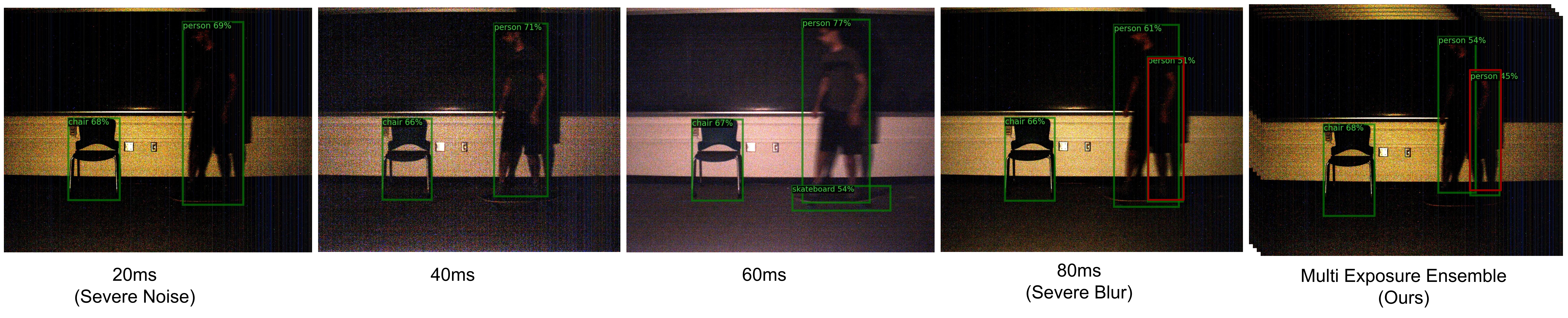}
\caption[]{\textbf{Object Detection Failure Cases on Real Captures:} Figure shows a failure case with the real captures where a single exposure (60ms) detects all three objects correctly whereas our model a detects false positive box and fails to detect skateboard object. Our model performs worse in cases when any single exposure has too much degradation. }
\label{fig:realfailuresupp}
\end{figure*}

\fi

\end{document}